\documentclass[journal]{IEEEtran}
\usepackage{amsmath,amsfonts}
\usepackage{algorithmic}
\usepackage{algorithm}
\usepackage{array}
\usepackage{textcomp}
\usepackage{stfloats}
\usepackage{url}
\usepackage{verbatim}
\usepackage{graphicx}
\usepackage{cite}
\usepackage{multirow}

\usepackage{booktabs}
\usepackage{subfigure}
\usepackage{hyperref}					
\hypersetup{colorlinks,
	linkcolor=red,%
	citecolor=green}

\begin{document}

\title{Deep Pulse-Coupled Neural Networks}

\author{Zexiang Yi, Jing Lian, Yunliang Qi, Zhaofei Yu,  Huajin Tang, Yide Ma and Jizhao Liu
\thanks{This work is supported by the Regional Project of the National Natural Science Foundation of China (Grant 82260364) and Natural Science Foundation of Gansu Province (Grants 21JR7RA510, 21JR7RA345). \textit{(Corresponding author: Jizhao Liu.)}
\quad

Zexiang Yi, Yunliang Qi, Yide Ma, and Jizhao Liu are with the School of Information Science and Engineering, Lanzhou University, Lanzhou 730000, China (e-mail: liujz@lzu.edu.cn).

Jing Lian is with the School of Electronics and Information Engineering, Lanzhou Jiaotong University, Lanzhou 730070, China.

Zhaofei Yu is with the Institute for Artificial Intelligence, Peking University, Beijing 100871, China.

Huajin Tang is with the College of Computer Science and Technology, Zhejiang University, Hangzhou 310027, China.
\quad 
}
}




\maketitle

\begin{abstract}
Spiking Neural Networks (SNNs) capture the information processing mechanism of the brain by taking advantage of spiking neurons, such as the Leaky Integrate-and-Fire (LIF) model neuron, which incorporates temporal dynamics and transmits information via discrete and asynchronous spikes. However, the simplified biological properties of LIF ignore the neuronal coupling and dendritic structure of real neurons, which limits the spatio-temporal dynamics of neurons and thus reduce the expressive power of the resulting SNNs. In this work, we leverage a more biologically plausible neural model with complex dynamics, i.e., a pulse-coupled neural network (PCNN), to improve the expressiveness and recognition performance of SNNs for vision tasks. The PCNN is a type of cortical model capable of emulating the complex neuronal activities in the primary visual cortex. We construct deep pulse-coupled neural networks (DPCNNs) by replacing commonly used LIF neurons in SNNs with PCNN neurons. The intra-coupling in existing PCNN models limits the coupling between neurons only within channels. To address this limitation, we propose inter-channel coupling, which allows neurons in different feature maps to interact with each other. Experimental results show that inter-channel coupling can efficiently boost performance with fewer neurons, synapses, and less training time compared to widening the networks. For instance, compared to the LIF-based SNN with wide VGG9, DPCNN with VGG9 uses only 50\%, 53\%, and 73\% of neurons, synapses, and training time, respectively. Furthermore, we propose receptive field and time dependent batch normalization (RFTD-BN) to speed up the convergence and performance of DPCNNs. We evaluate DPCNNs on four mainstream datasets, including static datasets such as MNIST, Fashion-MNIST, CIFAR-10, and the neuromorphic dataset N-MNIST. DPCNNs consistently outperform LIF-based SNNs, with notable performance achieved by the VGG9-based DPCNN, which achieves an accuracy of 94.2\% in the CIFAR-10 dataset, setting the state-of-the-art performance under this architecture.
\end{abstract}


\begin{IEEEkeywords}
Spiking neural networks, Pulse-coupled neural networks, Brain-like intelligence, Neuromorphic computing, Image recognition.
\end{IEEEkeywords}

\section{Introduction}

Spiking neural networks (SNNs) have emerged as a promising approach for developing bio-plausible and energy-efficient artificial intelligence (AI) systems \cite{roy_towards_2019}. Unlike traditional ANNs, SNNs represent the dynamic properties of neural systems and employ discrete spikes to encode, transmit, and process information. However, the majority of existing SNNs are based on the leaky integrate-and-fire (LIF) models\cite{li_brain_2023,yiLearningRulesSpiking2023}, which only capture a subset of the essential biological properties of real neurons, such as membrane potential integration and leakage. In fact, biological plausibility is a crucial consideration when creating brain-like intelligence, which is typically characterized by neuronal dynamics\cite{li_brain_2023}. On the one hand, the rich neural dynamics of biological neurons have great potential to advance the development of more general AI. On the other hand, utilizing bio-plausible mechanisms is beneficial in enhancing our understanding of the brain.

Driven by the significance of bio-plausibility, a question arises: \textit{can the performance of SNNs be efficiently improved by introducing bio-plausible mechanisms?} Several attempts have been made to answer this question. For example, inspired by the neural heterogeneity in the brain, 
Fang et al. \cite{fang_incorporating_2021} incorporated learnable time constants to improve the performance and convergence of SNNs. Sun et al. \cite{sunSynapse2023} and Wang et al. \cite{wang2022ltmd} introduced learnable thresholding to endow LIF neurons with the ability to optimize their threshold values. Rathi and Roy \cite{9556508} finetuned ANN-converted SNNs with a learnable membrane leak and firing threshold. Inspired by the lateral connections in biological neural networks, Cheng et al. \cite{cheng_lisnn_2020} proposed a neural network with lateral connections, obtaining stronger noise-robustness.
\par
\begin{figure*}[htbp]
	\centering
	\subfigure[LIF and PCNN neuron]{\includegraphics[width=0.375\textwidth]{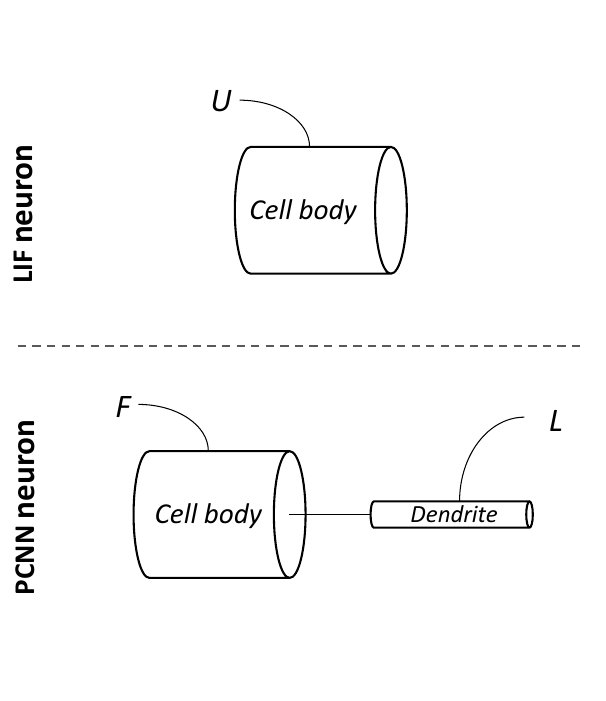}}
        \subfigure[Dynamics of the LIF neuron]{\includegraphics[width=0.55\textwidth]{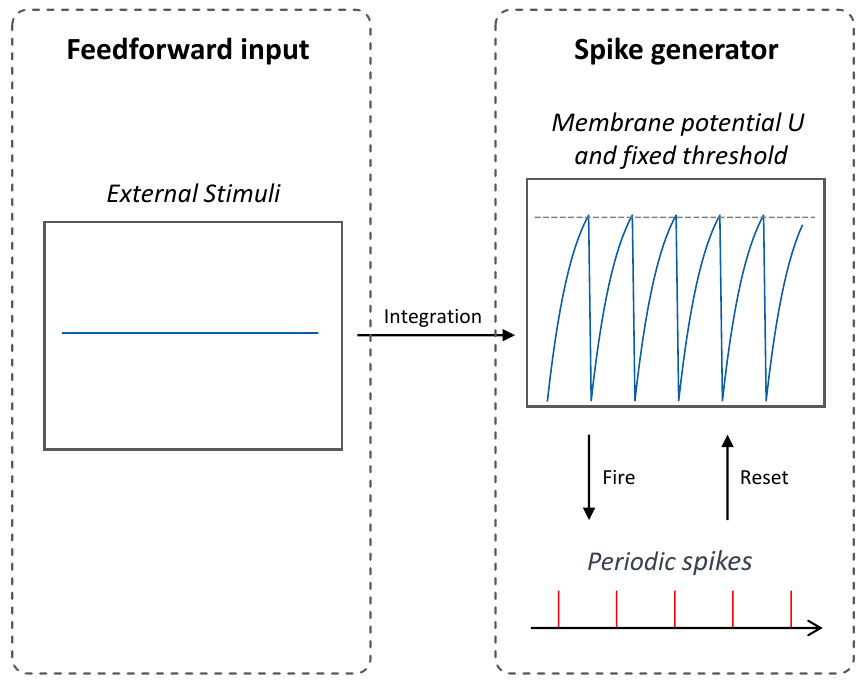}}
        \subfigure[Dynamics of the PCNN neuron]{\includegraphics[width=0.95\textwidth]{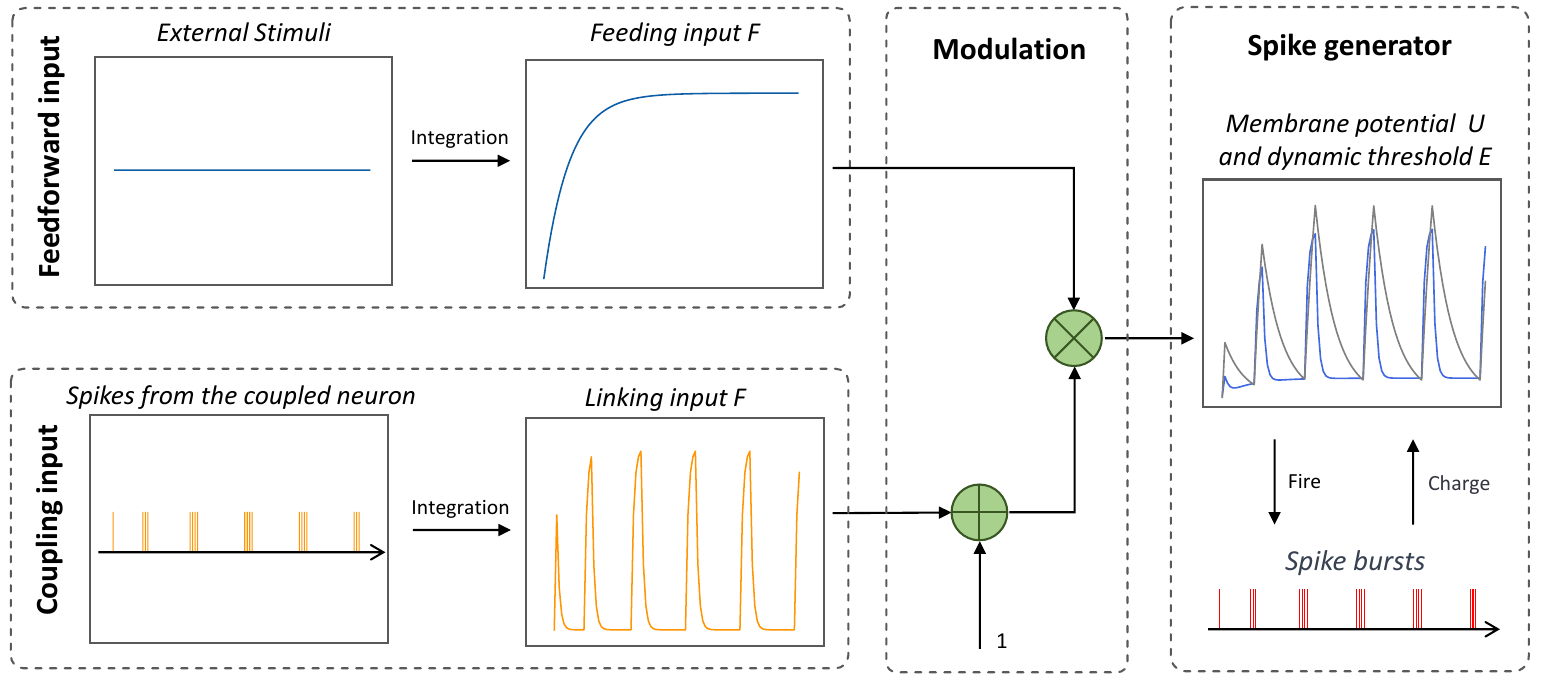}}
	\caption{Illustration of the LIF and PCNN neuron model and their dynamics: (a) In comparison to the LIF neuron, the PCNN neuron takes into account the dendritic structure, which enhances its computational capabilities. (b) The LIF neuron, when subjected to constant stimuli, exhibits periodic firing patterns. (c) Conversely, a pair of coupled PCNN neurons under constant stimuli can generate bursts of spikes, demonstrating their ability to exhibit synchronized activity and complex dynamics.}
	\label{fig:lif-pcnn}
\end{figure*}

These works focus on introducing bio-plausible mechanisms by modifying the LIF model. However, heuristic modifications of LIF may diverge from the dynamics of biological neural networks. We address this problem by utilizing direct biological experiment-derived models, namely pulse-coupled neural networks (PCNNs) \cite{eckhorn_feature_1990,johnson_pcnn_1999}. Fig. \ref{fig:lif-pcnn} illustrates the structures and dynamics of LIF and PCNN neurons. Unlike LIF neurons, PCNN neurons consider the coupling between neighboring neurons from linking input and modulation from dendritic computational dynamics \cite{johnson_pcnn_1999}.


The above observation motivated us to investigate deep pulse-coupled neural networks (DPCNNs), which are more biologically plausible and have more complex dynamic properties than existing SNNs. To introduce PCNN neurons into SNN architectures, we altered the feeding of PCNNs with learnable matrix-vector input. Besides, we found that the intra-channel coupling of PCNNs, which was also exploited by a previous SNN work \cite{cheng_lisnn_2020}, cannot improve the performance of DPCNNs (Tab. \ref{tab:intra-coupling}). Therefore, we introduced inter-channel coupling into DPCNNs, where neurons in different channels can communicate and coordinate with each other. Inter-channel coupling can enhance the network's expressive power, thus effectively improving the DPCNNs' performance. Additionally, to enable deeper architectures and further improve performance, we proposed receptive field and time-dependent batch normalization (RFTD-BN), which utilizes space- and time-varying BN parameters to capture the complex spatio-temporal dynamics of PCNNs.
\par
The main contributions of this work are summarized as follows:

\begin{itemize}
\item{We propose deep pulse-coupled neural networks (DPCNNs) by introducing PCNN neurons into SNN architectures and training them with spatio-temporal backpropagation (STBP). DPCNNs better represent the spatio-temporal dynamics of the visual cortex and thus have improved expressivity compared to existing SNNs (see Sec. \ref{sec:ablation}).} 
\item{We propose inter-channel coupling for DPCNNs, which enables neurons in different channels to communicate and cooperate with each other. We show that inter-channel coupling is essential to improving the performance of DPCNNs (Tab. \ref{tab:intra-coupling}). Additionally, to address the problem of DPCNNs' difficulty in convergence, we propose receptive field and time-dependent batch normalization (RFTD-BN), which can capture the complex spatio-temporal dynamics of PCNN (Fig. \ref{fig:rftd-bn}).}
\item{We demonstrate that incorporating more biologically plausible neural models with complex dynamics can enhance the performance of SNNs. For instance, VGG9-based DPCNN achieves an accuracy of 94.2\% on the CIFAR-10 dataset, which is the state-of-the-art performance under this architecture (Tab. \ref{tab:acc-comp}).}

\end{itemize}

\par
In our experiments, we analyze not only the effectiveness of inter-channel coupling and RFTD-BN but also the characteristics of DPCNNs, including the optimal coupling scale (Tab. \ref{tab:coupling-scale}), the necessity of multiplicative coupling (Tab. \ref{tab:coupling-add}), the impact of hyperparameters (see Sec. \ref{sec:hyperparameters}), and their robustness to noise (see Sec. \ref{sec:noise}).

\section{Related Works}
\subsection{Deep SNNs}
Generally, there are two strategies for achieving deep SNNs: direct training from scratch and conversion from ANN to SNN.
\subsubsection{Direct Training SNNs with LIF}
In this approach, spike-based learning algorithms are utilized to train SNNs from scratch. Among these methods, most prior studies leverage surrogate gradients for spatio-temporal backpropagation \cite{yiLearningRulesSpiking2023}. Zheng et al. \cite{zheng_going_2021} accomplished an impressive extension of directly-trained SNNs by leveraging the ResNet architecture \cite{he2016deep}, enabling them to reach a remarkable depth of 50 layers. To address the degradation issue in spiking ResNet, two SNN-oriented ResNet variants have been proposed: SEW-ResNet \cite{fang_deep_2021} and MS-ResNet \cite{hu_advancing_2021}. By introducing attention mechanisms in SNNs, Yao et al. \cite{yaoAttentionSpikingNeural2023} has made further strides in reducing the performance gap between ANNs and SNNs. However, even with such deep architectures, the performance of SNNs still lags behind ANNs. We argue that this limitation arises from the LIF type of neuronal dynamics, which hinders the expressive power of SNNs.
\subsubsection{ANN-SNN Conversion}
Leveraging the fact that the firing rate of IF models can approximate the activation value of ReLU functions, the ANN conversion approach first trains an ANN with constraints and then converts it into an SNN. The first ANN-SNN conversion method was proposed by Cao et al. \cite{cao_spiking_2015}, which was subsequently extended to deeper SNNs with a slight decrease in performance \cite{rueckauer_conversion_2017,han_rmp-snn_2020, hu_spiking_2021}. Recently, theoretical analysis of conversion errors has significantly reduced the inference latency \cite{li_free_2021,bu_optimal_2022}. The ANN conversion approach capitalizes on the superior performance of ANNs while avoiding the tremendous hardware overhead associated with the direct training of SNNs, enabling quicker implementation on low-power neuromorphic hardware. However, this approach is currently limited to IF neurons and static datasets, and it cannot exploit the temporal dynamics inherent in SNNs.
\subsection{PCNN Models}
There are two paths for the further development of PCNN models. The first path involves simplifying the PCNN model like spiking cortical model (SCM) \cite{zhan2009new} and simplified PCNN (SPCNN) \cite{chen2011new}. These modifications aim to enhance the image processing performance of PCNN while reducing computational costs.
\par
The second path focuses on enhancing the biological plausibility of PCNN. Huang et al. \cite{huang2016application} considered neural heterogeneity and proposed the Heterogeneous PCNN (HPCNN). Liu et al. \cite{liu_butterfly_2022} introduced the Continuous-Coupled Neural Network (CCNN), where the step function of neurons is replaced with a sigmoid function. With this modification, CCNN exhibits highly complex nonlinear dynamics when stimulated by periodic signals.
\par
Recently, researchers have started utilizing PCNN as a preprocessing tool for image recognition. Wang et al. \cite{wang2016leaf} utilized PCNN for feature extraction and combined it with SVM for leaf recognition. However, PCNN and its variants are limited to single-layer networks, and applications high-level vision tasks are rare. 
\section{Methods}
\subsection{Problem Statement and Challenges}
Existing SNNs commonly employ the LIF  model as the spiking neuron due to its simplicity and mathematical tractability \cite{li_brain_2023,yiLearningRulesSpiking2023}. By incorporating the LIF model into deep architectures borrowed from ANNs, such as VGG \cite{simonyan2014very} and ResNet \cite{he2016deep}, SNNs have demonstrated the ability to tackle complex tasks, including ImageNet classification. However, when considering the biological perspective, real neural networks perform tasks like perception, navigation, and decision-making within physiological constraints on network depth. Recent studies \cite{winding2023connectome} on insect brain connectomes have highlighted the significance of recurrent connections for computational capacity, a factor often overlooked by current SNNs employed in vision tasks.
\par
This motivation drives us to explore more biological models that can handle vision tasks with a relatively shallow architecture. The pulse-coupled neural network (PCNN) emerges as a promising candidate since it is a phenomenological model of the primary visual cortex \cite{johnson_pcnn_1999}. With its recurrent connections between neurons, PCNN can replicate the intricate neural activities observed in the primary visual cortex, including synchronous oscillations \cite{eckhorn_feature_1990}. By leveraging the spatio-temporal dynamics of PCNN, it becomes capable of performing low-level vision tasks such as image segmentation without the need for training \cite{zhan_computational_2017}. However, current PCNN models are constrained to single-layer architecture and lack synaptic plasticity, which limits their application in high-level vision tasks such as image recognition.
\par
Deep pulse-coupled neural networks (DPCNNs) are our attempt to address the challenges mentioned above. The remainder of this section will describe DPCNNs in detail.
\subsection{PCNN Neuron}

\begin{figure}[htbp]
	\centering
	\includegraphics[width=0.95\columnwidth]{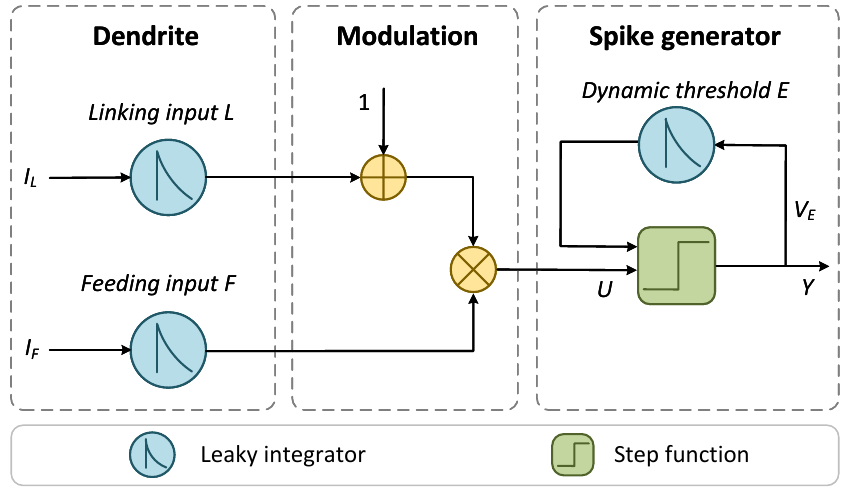}
	\caption{Diagram of the PCNN neuron.}
	\label{fig:pcnn_sch-en}
\end{figure}

The PCNN neuron is a two-compartmental neuron model \cite{johnson_pcnn_1999} and serves as the fundamental unit of a PCNN, as illustrated in Fig. \ref{fig:pcnn_sch-en}. It comprises three parts: the dendrite, the modulation, and the spike generator. Each part will be described in the the following. 


\par
The dendrite has two distinct input, the primary input termed feeding input $ F $, and the auxiliary input termed linking input $ L $. These two input channels integrate signals collected by their corresponding receptive fields, which can be represented as leaky integrators:
\begin{align}
	\tau_{F}\frac{dF}{dt} &=  -F + R_{F}I_{F} \label{eq:f}\\
	\tau_{L}\frac{dL}{dt} &=  -L + R_{L}I_{L} \label{eq:l} 
\end{align}	
where $ \tau_{F} $,  $ \tau_{L} $, $ I_{F} $, and $ I_{L} $ are the time constants and synaptic currents of the feeding and linking input, respectively.
\par
The linking input together with a constant offset term (+1), interact multiplicatively with the feeding input. The modulatory effect on the feeding input forms the membrane potential $U $ of the PCNN neruon:
\begin{equation}
	U = F\left(1 +  L\right)
\end{equation}
compared to the additive coupling, the modulatory coupling has the advantage that a neuron with no primary input can not be activated by the coupling input.
\par
Whenever the membrane potential reaches the threshold $ E $, the spike generator will emit a spike. Unlike LIF models resetting the membrane potential, the PCNN neuron feeds back to the threshold, which is another leaky integrator given by:
\begin{equation}
	\tau_{E}\frac{dE}{dt} =  -E + V_{E}S(t) \label{eq:e} 
\end{equation}
where $ S(t) $ denotes the output spike train of the PCNN neuron. $ \tau_{E} $ is the time constant and $ V_{E} $ is the amplitude gain. 
\par
In simulation, the Euler's method is often used to convert Eq. (\ref{eq:f}), (\ref{eq:l}), and (\ref{eq:e}) into discrete forms. Thus the PCNN neuron can be formalized as:
\begin{equation}
\label{eq:pcnnneuron1}
\left\{
\begin{aligned}
    F^{t} &= \left(1 - \frac{\Delta t}{\tau_{F}}\right) F^{t-1} + \frac{\Delta t}{\tau_{F}}R_{F}I^{t}_{F} \\
     L^{t} &= \left(1 - \frac{\Delta t}{\tau_{L}}\right) L^{t-1} + \frac{\Delta t}{\tau_{L}}R_{L}I^{t}_{L}  \\
     E^{t} &= \left(1 - \frac{\Delta t}{\tau_{E}}\right) E^{t-1} + \frac{\Delta t}{\tau_{E}}V_{E}Y^{t-1}  \\
     U^{t} &= F^{t}\left(1 + L^{t}\right) \\
     Y^{t} &= \Theta \left(U^{t} - E^{t}\right) 
\end{aligned}
\right.
\end{equation}
where $\Theta(x) = \begin{cases} 1, & x \ge 0\\ 0, & x < 0  \end{cases} $ is the step function, $t$ and $t-1$ denote the current time step and the previous time step, respectively, $\left(1 - \frac{\Delta t}{\tau_{F}}\right)$, $\left(1 - \frac{\Delta t}{\tau_{L}} \right)$ and $\left(1 - \frac{\Delta t}{\tau_{E}}\right)$ are the leak factors of the state variables $F$, $L$, and $E$, respectively. For the sake of brevity, we uses $\alpha_{F}$, $\alpha_{F}$, and $\alpha_{F}$ respectively to represent these three leak factors. Without loss of generality, we set $\frac{\Delta t}{\tau}R = 1$. Thus Eq. (\ref{eq:pcnnneuron1}) can be expressed as:
\begin{equation}
\label{eq:pcnnneuron2}
\left\{
\begin{aligned}
   F^{t} &= \alpha_{F} F^{t-1} + I^{t}_{F} \\
     L^{t} &= \alpha_{L} L^{t-1} + I^{t}_{L}  \\
     E^{t} &= \alpha_{E} E^{t-1} + V_{E}Y^{t-1}  \\
     U^{t} &= F^{t}\left(1 + L^{t}\right) \\
     Y^{t} &= \Theta \left(U^{t} - E^{t}\right) 
\end{aligned}
\right.
\end{equation}
\subsection{Building Blocks for DPCNN}
\begin{figure*}[t]
	\centering
	\includegraphics[width=0.75\textwidth]{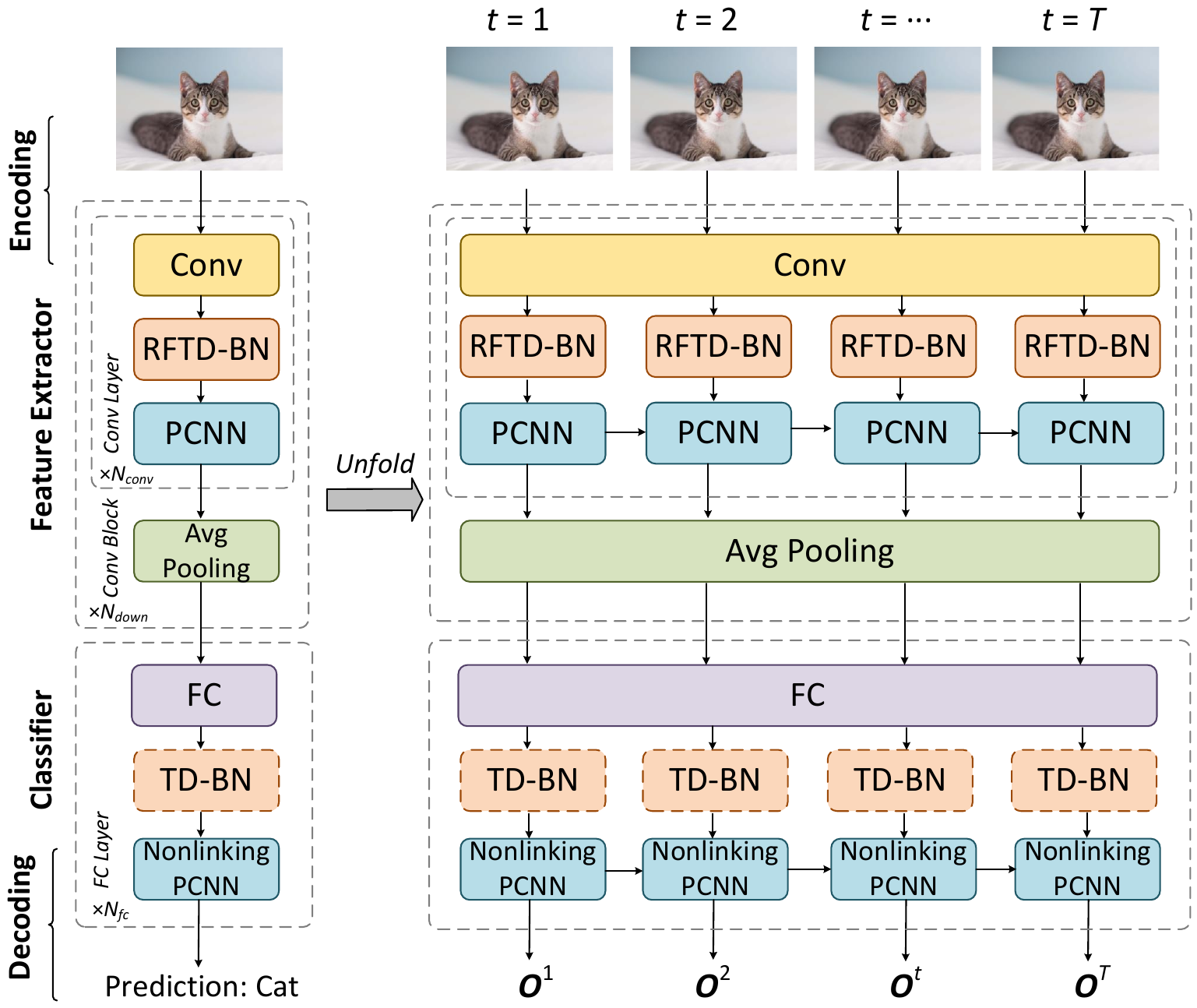}
	\caption{Diagram of the DPCNN model.}
	\label{fig:arch}
\end{figure*}
The DPCNN is depicted in Fig. \ref{fig:arch}. Functionally, DPCNN can be divided into four parts: neural encoding, feature extractor, classifier, and neural decoding. Neural encoding converts real-valued inputs into spikes, while neural decoding reverses this process. The feature extractor is comprised of multiple convolutional blocks, each consisting of multiple convolutional layers. The classifiers consist of fully connected layers. The structure of DPCNN is similar to conventional CNNs, except that the activation functions used in CNNs, such as ReLU, are replaced with PCNN and non-linking PCNN (PCNN without linking input) in DPCNN. To accommodate this change, we propose the receptive field and time-dependent batch normalization (RFTD-BN), and its simplified version, time-dependent batch normalization (TD-BN), which are used before the PCNN and nonlinking PCNN, respectively. DPCNN is a spatio-temporal network, which can be unfolded in the time domain. A detailed description of DPCNN will be presented in the remainder of this section.
\subsubsection{PCNN Layer}
PCNN is a neural network consisting of PCNN neurons that receive both external input and the influence of neighboring neurons. In the classical PCNN model\cite{wang_review_2010,zhan_computational_2017,lian_overview_2021}, each PCNN neuron represents a pixel, and the value of this pixel is input to the neuron as the synaptic current of the feeding input. To introduce learning into PCNN, we modify the feeding input to have an adaptive matrix-vector input. PCNN with convolutional connections can be expressed mathematically as follows:
\begin{equation}
\label{eq:pcnn}
\left\{
\begin{aligned}
    \boldsymbol{F}^{t,n} &= \alpha_{F} \boldsymbol{F}^{t-1,n} + \boldsymbol{W}^{n}\ast \boldsymbol{Y}^{t,n-1} \\
    \boldsymbol{L}^{t,n} &= \alpha_{L} \boldsymbol{L}^{t-1,n} + \boldsymbol{M}^{n}\ast \boldsymbol{Y}^{t-1,n} \\
    \boldsymbol{E}^{t,n} &= \alpha_{E} \boldsymbol{E}^{t-1,n} + V_{E}\boldsymbol{Y}^{t-1,n} \\
    \boldsymbol{U}^{t,n} &= \boldsymbol{F}^{t,n}\odot \left(1 + \boldsymbol{L}^{t,n}\right) \\
    \boldsymbol{Y}^{t,n} &= \Theta \left(\boldsymbol{U}^{t,n} - \boldsymbol{E}^{t,n}\right)
\end{aligned}
\right.
\end{equation}
where, $t$ and $n$ denote the current time step and the current PCNN layer, respectively, and $t-1$ and $n-1$ denote the previous time step and the previous PCNN, respectively. $\odot$ denotes element-wise multiplication, and $\ast$ denote convolution. The feeding input receives the output of the previous PCNN through the feedforward synaptic weights $\boldsymbol{W}^{n}$, resulting in the feeding synaptic current $\boldsymbol I_{F}=\boldsymbol{W}^{n}\ast \boldsymbol{Y}^{t,n-1}$. The linking input receives the coupled output of neighboring neurons through the coupling synaptic weights $\boldsymbol{M}^{n}$, resulting in the linking synaptic current $\boldsymbol I_{L}=\boldsymbol{M}^{n}\ast \boldsymbol{Y}^{t-1,n}$.
\par
\begin{figure*}[htbp]
	\centering
	\includegraphics[width=0.8\textwidth]{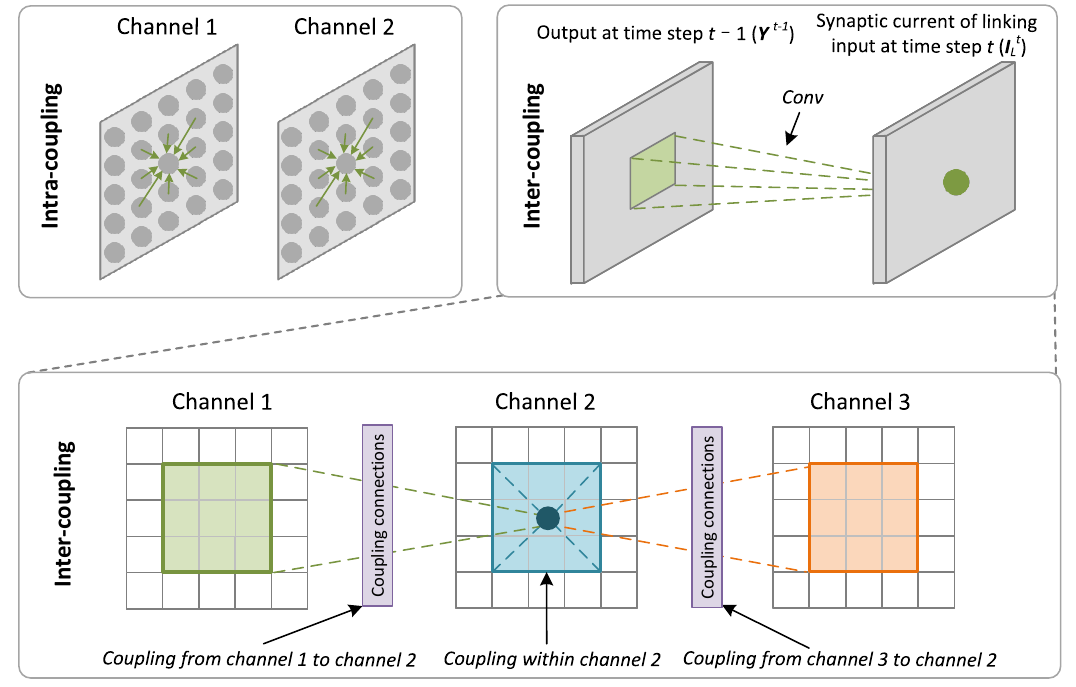}
	\caption{comparison between intra- and inter-coupling.}
	\label{fig:intra-coupling}
\end{figure*}

In previous works, coupling was restricted to intra-channel coupling \cite{lian_overview_2021} which ignores the interaction between channels. To this end, we propose inter-channel coupling, where neurons not only receive input from neighboring neurons in their own channel but also from neurons in other channels. Fig.\ref{fig:intra-coupling} illustrates the difference between intra-channel coupling and inter-channel coupling. In our work, we implement inter-channel coupling using multichannel convolution with a kernel size of $3\times3$, a stride of 1, and a padding of 1. It is worth noting that neurons also receive input from their own output, which can be interpreted as an auto-synapse.
\par
If the coupling between neurons is not considered, PCNN is reduced to nonlinking PCNN (PCNN without linking input):
\begin{equation}
\label{eq:npcnn}
\left\{
\begin{aligned}
    \boldsymbol{U}^{t,n} &= \boldsymbol{F}^{t,n} = \alpha_{F} \boldsymbol{F}^{t-1,n} + \boldsymbol{W}^{n}\ast \boldsymbol{Y}^{t,n-1} \\
\boldsymbol{E}^{t,n} &= \alpha_{E} \boldsymbol{E}^{t-1,n} + V_{E}\boldsymbol{Y}^{t-1,n}\\
\boldsymbol{Y}^{t,n} &= \Theta \left(\boldsymbol{U}^{t,n} - \boldsymbol{E}^{t,n}\right) 
\end{aligned}
\right.
\end{equation}
as shown in Fig. \ref{fig:arch} the nonlinking PCNN used in the classifier is a 1D networks with full connections.
\par
We initialize $\boldsymbol{F}^{t}$, $\boldsymbol{L}^{t}$ and $\boldsymbol{Y}^{t}$ with zeros. To avoid meaningless firing, $\boldsymbol{E}^{t}$ is initialized with a large value:
\begin{equation}
\boldsymbol{F}^{0}  = \boldsymbol{L}^{0}  = \boldsymbol{Y}^{0}  = 0 \text{ and }  \boldsymbol{E}^{0}  = \frac{{V}_{E}}{\alpha_{E}}  \label{eq:init}
\end{equation}
\par
PCNN and nonlinking PCNN can be implemented on PyTorch framework frameworks via Eq. (\ref{eq:pcnn}) to Eq. (\ref{eq:init}).
\subsubsection{RFTD-BN Layer}
\label{sec:rftd-bn}
Batch normalization (BN)\cite{ioffe2015batch} is a necessary technique for training deep networks, which can accelerate convergence and improve performance. BN is defined as follows:
\begin{equation}
    \mathrm{BN}(\boldsymbol{X}) = \boldsymbol{\gamma} \odot \frac{\boldsymbol{X} - {\boldsymbol{\mu}}_\mathcal{B}}{{\boldsymbol{\sigma}}_\mathcal{B}} + \boldsymbol{\beta} 
\end{equation}
where $\mathcal{B}$ denote a minibatch, $\boldsymbol{X}$ is an input sample, and $\gamma$ and $\beta$ are learnable scale and shit parameter, respectively. $\boldsymbol{\mu}_\mathcal{B}$ and $\boldsymbol{\sigma}_\mathcal{B}^2$ are the mean and variance of the minibatch, respectively, which can be calculated as follows:
\begin{align}
    \boldsymbol{\mu}_\mathcal{B} &= \frac{1}{|\mathcal{B}|} \sum_{\boldsymbol{X} \in \mathcal{B}} \boldsymbol{X} \\
    \boldsymbol{\sigma}_\mathcal{B}^2 &= \frac{1}{|\mathcal{B}|} \sum_{\boldsymbol{X} \in \mathcal{B}} (\boldsymbol{X}-\boldsymbol{\mu}_\mathcal{B})^2 + \epsilon
\end{align}
where $\epsilon$ is a small positive constant to guarantee numerical stability. During training, the mean and variance of the entire dataset are estimate through moving average for prediction.
\par
Recently, researchers try to incorporate BN to enhance the learning of SNNs. Fang et al. \cite{fang_incorporating_2021} uses shared BN paremeters ($\mu$, $\sigma$, $\gamma$, and $\beta$) at all time steps. Zheng et al. \cite{zheng_going_2021} extended BN to surpport spatial-temporal input and considered the neuronal threshold. Kim and Panda \cite{kim_revisiting_2021} argued that BNs that shared parameters on the time domain could not capture SNN dynamics, so they utilized separate sets of BN parameters for different time steps.
\par
However, due to the additional linking input, the existing BN methods cannot be applied to PCNN. In order to capture the complex spatio-temporal dynamics of PCNN, we proposes receptive field and time dependent batch normalization (RFTD-BN) which utilizes space- and time-varying BN parameters for the feeding and linking input at different time steps given by:
\begin{align}
    \hat{\boldsymbol{I}^{t}_{F}} = \mathrm{BN}_{F}^{t}(\boldsymbol{I}^{t}_{F}) &= \boldsymbol{\gamma}_{F}^{t} \odot \frac{\boldsymbol{I}^{t}_{F} - {\boldsymbol{\mu}}_{\mathcal{B},F}^{t}}{{\boldsymbol{\sigma}}_{\mathcal{B},F}^{t}} + \boldsymbol{\beta}^{t}_{F} \label{bn_f}\\ 
    \hat{\boldsymbol{I}^{t}_{L}} = \mathrm{BN}_{L}^{t}(\boldsymbol{I}^{t}_{L}) &= \boldsymbol{\gamma}_{L}^{t} \odot \frac{\boldsymbol{I}^{t}_{L} - {\boldsymbol{\mu}}_{\mathcal{B},L}^{t}}{{\boldsymbol{\sigma}}_{\mathcal{B},L}^{t}} + \boldsymbol{\beta}^{t}_{L} \label{bn_l}
\end{align}
where the subscripts $F$ and $L$ denote the feeding and linking input channel, respectively, and the superscript $t$ denotes the time step. As we can see, BN parameters depend on input channel (receptive field) and time. For nonlinking PCNN in the classifier of DPCNN, RFTD-BN is simplified to time dependent batch normalization (TD-BN). Additionally, the output nonlinking PCNN receive spikes from previous layer without TD-BN, we denote this practice with the dotted box of TD-BN in Fig. \ref{fig:arch}.
\subsubsection{Convolutional Block}
DPCNN employs multiple convolutional blocks to form a feature extractor that extracts features from input images, as depicted in Fig. \ref{fig:arch}. The variable $N_{block}$ represents the number of convolutional blocks, and $N_{conv}$ represents the number of convolutional layers in each convolution block. The convolutional block used in DPCNN is similar to the VGG block \cite{simonyan2014very}, utilizing a small kernel size of $3 \times 3$. In the convolutional block, a pooling layer is utilized to reduce the size of feature maps and the sensitivity of convolutional layers to location. Due to the binarity of spikes, average pooling is utilized to avoid significant information loss \cite{9556508} that may occur with max pooling, which is a common choice in ANNs. The extracted features are then classified using a classifier that consists of $N_{fc}$ fully connected layers.
\subsubsection{Neural Encoding and Decoding}
As shown in Fig. \ref{fig:arch}, direct encoding is utilized in DPCNN, which is similar to \cite{fang_incorporating_2021}, except that a PCNN layer is employed in the encoder instead of an SNN layer to convert real-valued signals into spikes.
\par
A nonlinking PCNN layer is considered as the neural decoder, which accumulates the incoming spikes without any leakage. The membrane potential at the last time step is set as the output of DPCNN, denoted as $\boldsymbol{y}_{net}$ and is determined by:
\begin{equation}
    \boldsymbol{y}_{net}= \textbf{\textit{U}}^{T,N} = \textbf{\textit{F}}^{T,N} =  \sum_{t=1}^{T} \textbf{\textit{I}}_{F}^{t,N} \label{output}
\end{equation}
\par
The prediction class is the index with the maximum value in $\boldsymbol{y}_{net}$, which can be obtained by calculating $\operatorname*{argmax} {y}_{net}^{i}$.
\subsection{Learning Rule for DPCNN}
The DPCNN can be unfolded as an RNN, as illustrated in Fig. \ref{fig:arch}, and train it using spatio-temporal backpropagation (STBP)\cite{wu_spatio-temporal_2018}. By utilizing Eq. (\ref{eq:pcnn}) and the loss function, we can derive the STBP formula for DPCNN, which we will describe in detail in the followings.
\subsubsection{Loss Function}
We employ the temporal efficient training cross-entropy\cite{deng_temporal_2022} as the loss function of DPCNN, which can help smooth the loss landscape and improve performance \cite{deng_temporal_2022}. It is described as:
\begin{align}
\mathcal{L} &= \frac{1}{T} \sum_{t=1}^{T}  \mathcal{L}_{CE} \left (  \operatorname*{softmax} (\boldsymbol{O}^{t}),\boldsymbol{y} \right ) \nonumber  \\
&= \frac{1}{T} \sum_{t=1}^{T}  \mathcal{L}_{CE} \left (  \hat{\boldsymbol{O}^{t}},\boldsymbol{y} \right ) \nonumber \\
&= -\frac{1}{T} \sum_{t=1}^{T} \sum_{i}^{}{y}_{i} \log_{}{\hat{{O}^{t}_{i}}}  \label{loss}
\end{align}
where $\mathcal{L}_{CE}$ denotes cross-entropy loss, $\boldsymbol{y}$ represents the one-hot coding of the target label, $\boldsymbol{O}^{t}$ represents the synaptic current at time step $t$ of the output nonlinking PCNN. The $\operatorname*{softmax}$ function transforms $\boldsymbol{O}^{t}$ into a probability vector $\hat{\boldsymbol{O}^{t}}$, with each element $\hat{{O}^{t}_{i}} = \frac{\exp({O}^{t}_{i})}{ \sum{j}^{} \exp({O}^{t}_{j})}$.
\subsubsection{STBP in DPCNN}
Spatial-temporal backpropagation (STBP) is a direct training method that treats SNNs as RNNs. It calculates gradients for weight updating through the backpropagation through time (BPTT) algorithm \cite{wu_spatio-temporal_2018}. To overcome the non-differentiability of the step function, STBP uses a surrogate gradient for gradient estimation.
\par
\begin{figure}[htbp]
	\centering
	\includegraphics[width=\columnwidth]{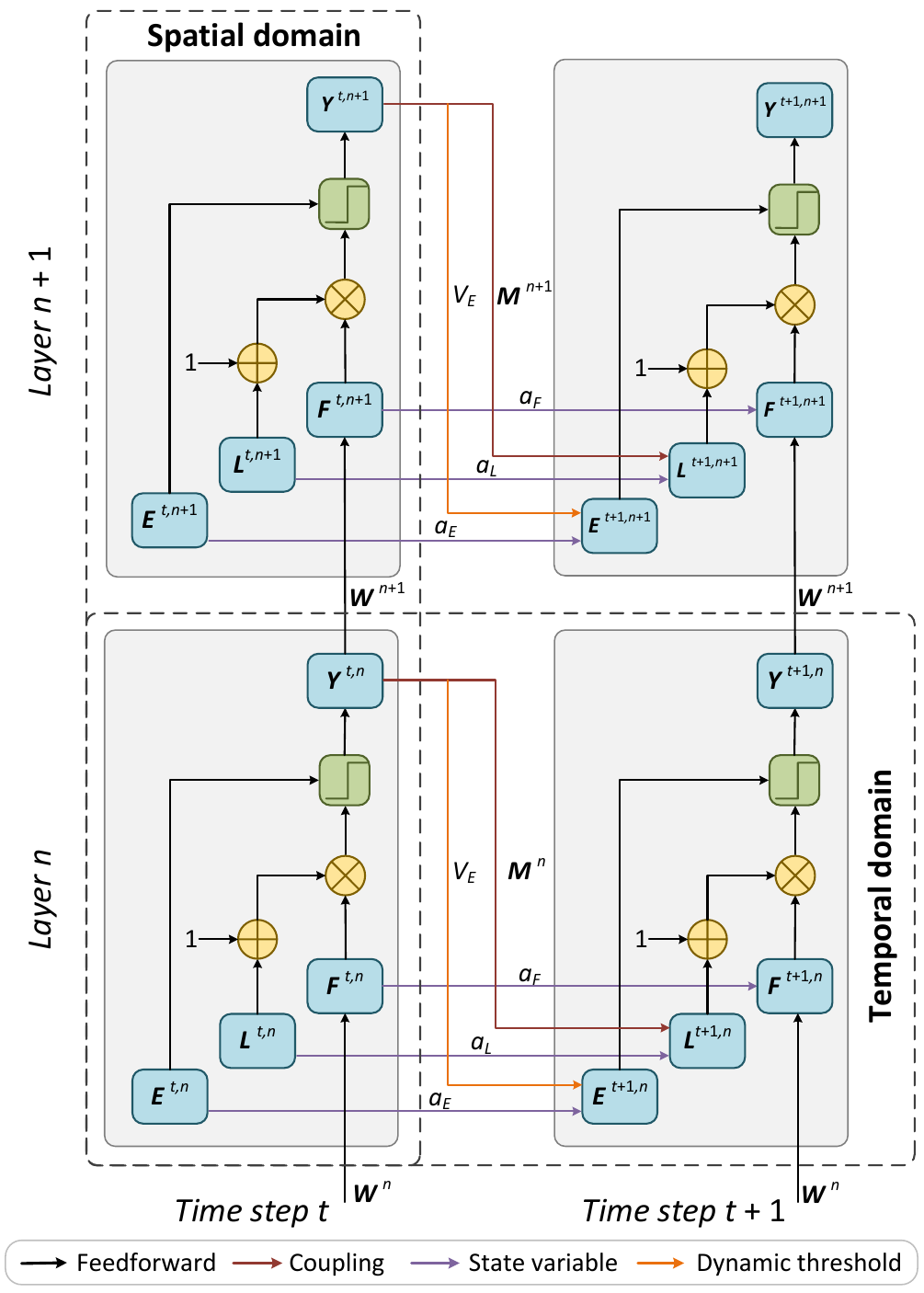}
	\caption{Computational gragh of DPCNN.  Information propagates in both the spatial and temporal domains. The backpropagation path is not included in the diagram for the sake of brevity.}
	\label{fig:forward}
\end{figure}

The computational graph of the RNN-like expression of DPCNN is illustrated in Fig. \ref{fig:forward}. It can be observed that information not only propagates over the spatial domain (from layer $n$ to layer $n+1$) through feedforward connections but also through state variables of neurons and coupling connections in the temporal domain (from time step $t$ to time step $t+1$).
\par
The feeding weights $\boldsymbol{W}^{n}$ indirectly impact the feeding input $\boldsymbol{F}^{t,n}$ by influencing the feeding synaptic current $\boldsymbol{I}^{t,n}_{F}$, which ultimately changes the loss. Additionally, the output layer takes the feeding synaptic current $\boldsymbol{I}^{t,n}{F}$ as its output $\boldsymbol{O}^{t}$. By applying the chain rule, in combination with Eq. (\ref{eq:pcnn}) and (\ref{loss}), we can obtain the gradient of the loss function with respect to $\boldsymbol{W}^{n}$:
\begin{align}
\frac{\partial \mathcal{L}}{\partial \boldsymbol{W}^{n}}  &= \sum_{t  = 1}^{T} \frac{\partial \mathcal{L}}{\partial \boldsymbol{I}^{t,n}_{F}}\frac{\partial \boldsymbol{I}^{t,n}_{F}}{\partial \boldsymbol{W}^{t,n}} \nonumber \\
&= \begin{cases} 
  \sum_{t  = 1}^{T} \frac{\partial \mathcal{L}}{\partial \boldsymbol{F}^{t,n}} \frac{\partial \boldsymbol{F}^{t,n}}{\partial \boldsymbol{I}^{t,n}_{F}} \frac{\partial \boldsymbol{I}^{t,n}_{F}}{\partial \boldsymbol{W}^{t,n}}, & n<N \\
  \sum_{t  = 1}^{T} \frac{\partial \mathcal{L}}{\partial \boldsymbol{O}^{t}}  \frac{\partial \boldsymbol{O}^{t}}{\partial \boldsymbol{W}^{N}}, & n=N
\end{cases} \nonumber  \\
&= \begin{cases} 
  \sum_{t  = 1}^{T} \frac{\partial \mathcal{L}}{\partial \boldsymbol{F}^{t,n}} {\boldsymbol{Y}^{t,n-1}}^\mathsf{T}, & n<N \\
  \sum_{t  = 1}^{T} \left [ \operatorname*{softmax} (\boldsymbol{O}^{t})-\boldsymbol{y} \right ]  {\boldsymbol{Y}^{t,N-1}}^\mathsf{T}, & n=N
\end{cases} \label{eq:gradw}
\end{align}
\par
Similarly, the linking weights $\boldsymbol{M}^{n}$ only indirectly affect the linking input $\boldsymbol{L}^{t,n}$ by altering the input current $\boldsymbol{I}^{t,n}_{L}$, ultimately leading to a change in the loss. Applying the chain rule and combining it with Eq. (\ref{eq:pcnn}), we obtain:
\begin{align}
\frac{\partial \mathcal{L}}{\partial \boldsymbol{M}^{n}}  &= \sum_{t = 1}^{T} \frac{\partial \mathcal{L}}{\partial \boldsymbol{I}^{t,n}_{L}}\frac{\partial \boldsymbol{I}^{t,n}_{L}}{\partial \boldsymbol{M}^{t,n}} \nonumber \\
&= 
  \sum_{t  = 1}^{T} \frac{\partial \mathcal{L}}{\partial \boldsymbol{L}^{t,n}} \frac{\partial \boldsymbol{L}^{t,n}}{\partial \boldsymbol{I}^{t,n}_{L}} \frac{\partial \boldsymbol{I}^{t,n}_{L}}{\partial \boldsymbol{M}^{t,n}} \nonumber \\
&= 
  \sum_{t  = 1}^{T} \frac{\partial \mathcal{L}}{\partial \boldsymbol{L}^{t,n}} {\boldsymbol{Y}^{t-1,n}}^\mathsf{T} \label{eq:gradm}
\end{align} 
From Eq. (\ref{eq:gradw}) and (\ref{eq:gradm}), it can be seen that the key to computing $\frac{\partial \mathcal{L}}{\partial \boldsymbol{W}^{n}}$ and $\frac{\partial \mathcal{L}}{\partial \boldsymbol{M}^{n}}$ is to calculate $\frac{\partial \mathcal{L}}{\partial \boldsymbol{F}^{t,n}}$ and $\frac{\partial \mathcal{L}}{\partial \boldsymbol{L}^{t,n}}$. To simplify notation, we introduces two symbols:
\begin{equation}
\boldsymbol{\delta} ^{t,n}_{F} \equiv 
  \frac{\partial \mathcal{L}}{\partial \boldsymbol{F}^{t,n}} \text{ and }
\boldsymbol{\delta} ^{t,n}_{L} \equiv \frac{\partial \mathcal{L}}{\partial \boldsymbol{L}^{t,n}} \label{eq:deltal}
\end{equation}
\par
Note that when $n=N$, $\boldsymbol{\delta} ^{t,N}_{F} \equiv \operatorname*{softmax} (\boldsymbol{O}^{t})-\boldsymbol{y}$. Similar to the backpropagation algorithm in ANN, $\boldsymbol{\delta} ^{t,n}{F}$ and $\boldsymbol{\delta} ^{t,n}_{F}$ can be regarded as errors. Therefore, the gradient of the loss with respect to the weight is equal to the input of the neuron multiplied by the error.
\par
For $\boldsymbol{\delta} ^{t,n}_{F}$, it can be seen from Fig. \ref{fig:forward} that within the neuron, $\boldsymbol{F}^{t,n}$ affects the output $\boldsymbol{Y}^{t,n}$ by influencing the membrane potential $\boldsymbol{U}^{t,n}$, ultimately affecting the loss. Meanwhile, in the temporal domain, $\boldsymbol{F}^{t,n}$ decays and is transmitted to the next time step. According to Eq. (\ref{eq:pcnn}) and the chain rule, it can be obtained that:
\begin{align}
\boldsymbol{\delta}^{t,n}_{F}   & = \frac{\partial \mathcal{L}}{\partial \boldsymbol{Y}^{t,n}} \frac{\partial \boldsymbol{Y}^{t,n}}{\partial \boldsymbol{U}^{t,n}} \frac{\partial \boldsymbol{U}^{t,n}}{\partial \boldsymbol{F}^{t,n}} +\frac{\partial \mathcal{L}}{\partial \boldsymbol{F}^{t+1,n}} \frac{\partial \boldsymbol{F}^{t+1,n}}{\partial \boldsymbol{F}^{t,n}} \nonumber \\
 & = 
 \begin{cases}
  \frac{\partial \mathcal{L}}{\partial \boldsymbol{Y}^{t,n}} \odot \Theta'(\boldsymbol{U}^{t,n}-\boldsymbol{E}^{t,n}) \odot (1+\boldsymbol{L}^{t,n }) \\+
  \underbrace{\boldsymbol{\delta} ^{t+1,n}_{F}}_{temporal} \alpha _{F}, & t<T \\
  \frac{\partial \mathcal{L}}{\partial \boldsymbol{Y}^{T,n}} \odot \Theta'(\boldsymbol{U}^{T,n}-\boldsymbol{E}^{T,n}) \odot (1+\boldsymbol{L}^{T,n }), & t=T
 \end{cases} \label{eq:error-f}
\end{align}
where $\odot$ denotes element-wise multiplication.
\par
We can use a similar approach to calculate $\boldsymbol{\delta} ^{t,n}_{L}$:
\begin{align}
\boldsymbol{\delta}^{t,n}_{L}  & = \frac{\partial \mathcal{L}}{\partial \boldsymbol{Y}^{t,n}} \frac{\partial \boldsymbol{Y}^{t,n}}{\partial \boldsymbol{U}^{t,n}} \frac{\partial \boldsymbol{U}^{t,n}}{\partial \boldsymbol{L}^{t,n}} +\frac{\partial \mathcal{L}}{\partial \boldsymbol{L}^{t+1,n}} \frac{\partial \boldsymbol{L}^{t+1,n}}{\partial \boldsymbol{L}^{t,n}} \nonumber \\
 & = 
\begin{cases}
 \frac{\partial \mathcal{L}}{\partial \boldsymbol{Y}^{t,n}} \odot \Theta'(\boldsymbol{U}^{t,n}-\boldsymbol{E}^{t,n}) \odot  \boldsymbol{F}^{t,n} \\ + 
  \underbrace{\boldsymbol{\delta} ^{t+1,n}_{L}}_{temporal} \alpha _{L}, &t<T   \\
  \frac{\partial \mathcal{L}}{\partial \boldsymbol{Y}^{T,n}} \odot \Theta'(\boldsymbol{U}^{T,n}-\boldsymbol{E}^{T,n}) \odot  \boldsymbol{F}^{T,n}, & t=T
\end{cases} \label{eq:error-l}
\end{align}
\par
Calculating $\frac{\partial \mathcal{L}}{\partial \boldsymbol{Y}^{t,n}}$ is critical. $\boldsymbol{Y}^{t,n}$ influences the loss $\mathcal{L}$ through the feeding weights $\boldsymbol{W}^{n+1}$, the linking weights $\boldsymbol{M}^{n}$, and the dynamic threshold $\boldsymbol{Y}^{t+1,n}$. Therefore, we have:
\begin{align}
\frac{\partial \mathcal{L}}{\partial \boldsymbol{Y}^{t,n}} & = \frac{\partial \mathcal{L}}{\partial \boldsymbol{F}^{t,n+1}} \frac{\partial \boldsymbol{F}^{t,n+1}}{\partial \boldsymbol{Y}^{t,n}} +\frac{\partial \mathcal{L}}{\partial \boldsymbol{L}^{t+1,n}} \frac{\partial \boldsymbol{L}^{t+1,n}}{\partial \boldsymbol{Y}^{t,n}} \nonumber\\ & \quad + \frac{\partial \mathcal{L}}{\partial \boldsymbol{E}^{t+1,n}} \frac{\partial \boldsymbol{E}^{t+1,n}}{\partial \boldsymbol{Y}^{t,n}} \nonumber\\
&= \begin{cases}
\underbrace{\boldsymbol{\delta}^{t,n+1}_{F}}_{spatial} \boldsymbol{W}^{n+1} +  \underbrace{\boldsymbol{\delta} ^{t+1,n}_{L}}_{temporal} \boldsymbol{M}^{n}\\  + \underbrace{\boldsymbol{\delta}^{t+1,n}_{E}}_{temporal} V_{E}, &t<T \\
\underbrace{\boldsymbol{\delta}^{t,n+1}_{F}}_{spatial}  \boldsymbol{W}^{n+1}, &t=T
\end{cases} \label{eq:gradient-y}
\end{align}
where $\boldsymbol{\delta} ^{t+1,n}_{E} \equiv
\frac{\partial \mathcal{L}}{\partial \boldsymbol{E}^{t+1,n}}$. To derive an expression for $\boldsymbol{\delta} ^{t,n}{E}$, we can apply the chain rule and obtain:
\begin{align}
\boldsymbol{\delta}^{t,n}_{E}    & = \frac{\partial \mathcal{L}}{\partial \boldsymbol{Y}^{t,n}} \frac{\partial \boldsymbol{Y}^{t,n}}{\partial \boldsymbol{E}^{t,n}} +\frac{\partial \mathcal{L}}{\partial \boldsymbol{E}^{t+1,n}} \frac{\partial \boldsymbol{E}^{t+1,n}}{\partial \boldsymbol{E}^{t,n}} \nonumber \\
 & = 
\begin{cases}
 -\frac{\partial \mathcal{L}}{\partial \boldsymbol{Y}^{t,n}} \odot \Theta'(\boldsymbol{U}^{t,n}-\boldsymbol{E}^{t,n})  + 
  \underbrace{\boldsymbol{\delta} ^{t+1,n}_{E}}_{temporal} \alpha _{E}, &t<T   \\
  -\frac{\partial \mathcal{L}}{\partial \boldsymbol{Y}^{T,n}} \odot \Theta'(\boldsymbol{U}^{T,n}-\boldsymbol{E}^{T,n}) , & t=T
\end{cases} \label{eq:error-e}
\end{align}
\par
By applying Eq. (\ref{eq:error-f}) to (\ref{eq:error-e}), we can compute the gradients recursively. 
\par
In the above equations, the derivative of the spike with respect to the membrane potential needs to be computed. However, $\Theta'(x)$ is zero everywhere except at $x=0$, which halts the flow of gradients. In this work, we choose the surrogate derivative given by:
\begin{equation}
  \Theta'(x) = \frac{1}{1+(\pi x)^{2} }  
\end{equation}
\par
The overall training algorithm is shown in Alg. \ref{alg:stbp}.


\begin{algorithm}[H]
    \caption{The overall training algorithm for DPCNN\label{alg:stbp}}
    \begin{algorithmic}[1]
      \REQUIRE input $\boldsymbol{X}=\left \{  \boldsymbol{X}^{1},\boldsymbol{X}^{2},...,\boldsymbol{X}^{T}\right \}$, learning rate $\eta$, number of time steps $T$
      \STATE initialize $\boldsymbol{W}^{n}$ and $\boldsymbol{M}^{n}$, state variables $\boldsymbol{E}^{0,n}  = \frac{{V}_{E}}{\alpha_{E}}$, $\boldsymbol{F}^{0,n}  = \boldsymbol{L}^{0,n}  = \boldsymbol{Y}^{0,n}  = 0 $
      \STATE create an empty list $\boldsymbol{O}=\left \{\right \} $
      \FOR{$t = 1,2,...,T$}
      \STATE feed $\boldsymbol{X}^{t}$ to network, get output $\boldsymbol{O}^{t}$
      \STATE append $\boldsymbol{O}^{t}$ to $\boldsymbol{O}$
      \ENDFOR
      \STATE calculate loss $\mathcal{L}(\boldsymbol{O},\boldsymbol{y})$
      \STATE calculate gradients $\frac{\partial \mathcal{L}}{\partial \boldsymbol{W}^{n}}, \frac{\partial \mathcal{L}}{\partial \boldsymbol{M}^{n}} \leftarrow Autograd$
      \STATE update weights$\boldsymbol{W}^{n}  \leftarrow \boldsymbol{W}^{n}-\eta\frac{\partial \mathcal{L}}{\partial \boldsymbol{W}^{n}}, \boldsymbol{M}^{n}  \leftarrow \boldsymbol{M}^{n}-\eta\frac{\partial \mathcal{L}}{\partial \boldsymbol{M}^{n}}$
    \end{algorithmic}
\end{algorithm}

\section{Experiments}
\subsection{Datasets and Preprocessing}
\subsubsection{MNIST}
MNIST dataset \cite{lecun_gradient-based_1998} is a handwritten digit dataset containing 10 classes of digits from 0 to 9. Each sample is a grayscale image with a size of 28x28. The dataset has a total of 70,000 samples, of which 60,000 are used for training and 10,000 for testing.
\subsubsection{N-MNIST}
The N-MNIST dataset \cite{orchard_converting_2015} is a neuromorphic version of the MNIST dataset that represents information through asynchronous events. Each sample has a resolution of 34x34. However, due to the large number of events, the dataset requires event-to-frame integrating methods. In this work, we use the SpikingJelly library \cite{fang2020spikingjelly} to split the events into 10 slices with an almost equal number of events in each slice, and then integrate them into frames.
\subsubsection{Fashion-MNIST}
Similar to the MISNT dataset, Fashion-MNIST\cite{xiao2017fashion} is a dataset of clothing images, consisting of 10 categories. It includes 60,000 training samples and 10,000 testing samples, with each image represented as a grayscale 28x28 pixel matrix.
\subsubsection{CIFAR-10}
The CIFAR10 dataset \cite{krizhevsky_cifar-10_2014} comprises 60,000 natural images of size 32x32, spread across 10 classes, with 6,000 images for each class. The training set contains 50,000 images, while the test set contains 10,000 images. In this paper, We perform data normalization, random cropping, random horizontal flipping, and cutout.

\subsection{Networks and Hyperparameters}

The network architectures for different datasets are shown in Table \ref{tab:arch}. \textit{32C} represents a convolutional layer with 32 channels and a kernel size of 3x3. \textit{P2} represents a pooling layer with a kernel size of 2x2.
\par
To ensure maximum reproducibility, we use same random seed in all experiments during training. We use the Adam \cite{kingma2014adam} optimizer with the CosineAnnealingLR \cite{loshchilov2016sgdr} learning rate scheduler. The value of $T_{max}$ is set to the number of training epochs. The hyperparameters used in our experiments are shown in Table \ref{tab:hyperpara}.
\begin{table}[htbp]
\renewcommand{\arraystretch}{1.2}
\centering
\caption{Network architectures for different datasets}
\label{tab:arch}
\begin{tabular}{ccc}
\toprule[1.2pt]
Dataset & Network & Architecture \\ 
\midrule[0.8pt]
MNIST & \multirow{3}{*}{MNISTNet} & \multirow{3}{*}{32C3-P2-32C3-P2-128-10} \\
N-MNIST &  &  \\
Fashion-MNIST &  &  \\ 
CIFAR-10 & VGG9 & \begin{tabular}[c]{@{}l@{}}64C3-64C3-P2-128C3-128C3-P2-\\ 256C3-256C3-256C3-P2-1024-10\end{tabular} \\
\bottomrule[1.2pt]
\end{tabular}
\end{table}

\begin{table}[htbp]
\renewcommand{\arraystretch}{1.2}
\centering
\caption{Hyperparameter settings for different datasets}
\label{tab:hyperpara}
\begin{tabular}{ccccc}
\toprule[1.2pt]
Hyperparameter          & MNIST & N-MNIST & Fashion-MNIST & CIFAR-10 \\
\midrule[0.8pt]
Time step ($T$) & 4 & 10 & 6 & 2/4/6/8 \\
Learning rate & 0.001 & 0.001 & 0.001 & 0.001 \\
Batch size & 50 & 50 & 50 & 50 \\
Epoch & 50 & 50 & 50 & 200 \\
$\alpha_{F}$ & 0.5 & 0.5 & 0.5 & 0.5 \\
$\alpha_{L}$ & 0.5 & 0.5 & 0.5 & 0.5 \\
$\alpha_{E}$ & 0.7 & 0.7 & 0.7 & 0.7 \\
$V_{E}$ & 1 & 1 & 1 & 1 \\
\bottomrule[1.2pt]
\end{tabular}
\end{table}


\subsection{Comparison with the State-of-the-Art}
We compares DPCNN with SOTA methods on static MNIST, Fashion-MNIST, and CIFAR-10 datasets, and neuromorphic N-MNIST dataset as shown in Tab. \ref{tab:acc-comp}. The results of the comparison will be discussed below.
\par
\textbf{MNIST}: PCNN outperforms LISNN with fewer time steps under the same network architecture (MNISTNet). DPCNN has a slightly lower accuracy of 0.12\% compared to the current SOTA method PLIF, but PLIF uses a much larger network size than DPCNN's MNISTNet. Additionally, PLIF uses twice the time step used by DPCNN.

\par
\textbf{N-MNIST}: The performance of DPCNN is only slightly lower than that of PLIF. Similar to the case in MNIST, DPCNN uses a smaller network size. Moreover, compared to LISNN with the same network configuration, DPCNN achieves a nearly 0.1\% improvement while using fewer time steps. 

\par
\textbf{Fashion-MNIST}: DPCNN achieves SOTA performance on this dataset, with an accuracy of 94.53\% in only 6 time steps. Additionally, DPCNN outperforms LISNN with the same network configuration by more than 2\%, demonstrating the superiority of DPCNN in terms of performance.

\textbf{CIFAR-10}: DPCNN is achieving SOTA performance with a 94.20\% accuracy at the same or similar network scale. For instance, compared with TEBN under the same network architecture (VGG9) and time step ($T=4$), DPCNN outperforms it by more than 1\%. Although DPCNN has lower accuracy than QCFS, which is based on ANN-SNN conversion, QCFS only applies to static images, and when under ultra latency ($T=2$), DPCNN has a significant advantage over QCFS. Additionally, DPCNN achieves comparable accuracy to the TET-trained ResNet-19. Overall, DPCNN achieves the best performance with similar network scale, and comparable accuracy to deep networks like ResNet-19 with fewer layers.

\begin{table*}[htbp]
\caption{Performance comparison betweeen DPCNN and existing methods on MNIST, N-MNIST, Fashion-MNIST and CIFAR-10. Bold texts indicate the proposed method\label{tab:acc-comp}}
\renewcommand{\arraystretch}{1.2}
\centering
\begin{tabular}{cccccc}
\toprule[1.2pt]
Dataset & Model & Learning Rule & Network & $T$ & Accuracy\\
\midrule[0.8pt]
\multirow{4}{*}{MNIST}  & STBP\cite{wu_spatio-temporal_2018} & Direct training & 15C5-P2-40C5-P2-300-10 & >100 & 99.42\% \\
& LISNN\cite{cheng_lisnn_2020} & Direct training & MNISTNet & 20 & 99.50\% \\
 & TSSL-BP\cite{NEURIPS2020_8bdb5058} & Direct training & 15C5-P2-40C5-P2-300-10 & 5 & 99.53\% \\
 & PLIF\cite{fang_incorporating_2021} & Direct training & 128C3-P2-128C3-P2-2048-100 & 8 & 99.72\% \\
 & \textbf{DPCNN} & Direct training & MNISTNet & \textbf{4} & \textbf{99.60\%} \\ \cline{1-6}
\multirow{4}{*}{N-MNIST} & LISNN\cite{cheng_lisnn_2020} & Direct training & MNISTNet & 20 & 99.45\% \\
 & TSSL-BP\cite{NEURIPS2020_8bdb5058} & Direct training & 12C5-P2-64C5-P2-10 & 30 & 99.28\% \\
 & PLIF\cite{fang_incorporating_2021} & Direct training & 128C3-P2-128C3-P2-2048-100 & 10 & 99.61\% \\
 & \textbf{DPCNN} & Direct training & MNISTNet & \textbf{10} & \textbf{99.54\%} \\ \cline{1-6}
\multirow{4}{*}{Fashion-MNIST} & LISNN\cite{cheng_lisnn_2020} & Direct training & MNISTNet & 20 & 92.07\% \\
 & TSSL-BP\cite{NEURIPS2020_8bdb5058} & Direct training & 32C5-P2-64C5-P2-1024-10 & 5 & 92.83\% \\
 & PLIF\cite{fang_incorporating_2021} & Direct training & 128C3-P2-128C3-P2-2048-100 & 8 & 94.38\% \\
 & \textbf{DPCNN} & Direct training & MNISTNet & \textbf{6} & \textbf{94.53\%} \\ \cline{1-6}
\multirow{13}{*}{CIFAR-10} & TSSL-BP\cite{NEURIPS2020_8bdb5058} & Direct training & CIFARNet & 5 & 91.41\% \\
 & PLIF\cite{fang_incorporating_2021} & Direct training & 9-layer CNN & 8 & 93.50\% \\
 & BNTT\cite{kim_revisiting_2021} & Direct training & VGG9 & 25 & 90.50\% \\
 & TEBN\cite{duan2022temporal} & Direct training & VGG9 & 4 & 92.80\% \\
 & DIET-SNN\cite{9556508} & ANN-SNN Conversion & VGG16 & 10 & 93.44\% \\
 & opt. mem.\cite{bu_optimized_2022} & ANN-SNN Conversion & VGG16 & 16 & 93.38\% \\   \cline{2-6}
 & \multirow{2}{*}{QCFS\cite{bu_optimal_2022}} & \multirow{2}{*}{ANN-SNN Conversion} & \multirow{2}{*}{VGG16} & 4 & 93.96\% \\
 &  &  &  & 2 & 91.18\% \\   \cline{2-6}
 & \multirow{3}{*}{TET\cite{deng_temporal_2022}} & \multirow{3}{*}{Direct training} & \multirow{3}{*}{ResNet-19} & 6 & 94.50 \\
 &  &  &  & 4 & 94.44 \\
 &  &  &  & 2 & 94.16 \\ \cline{2-6}
 & \multirow{4}{*}{\textbf{DPCNN}} & \multirow{4}{*}{Direct training} & \multirow{4}{*}{VGG9} & \textbf{8} & \textbf{94.20\%} \\   
 &  &  &  & \textbf{6} & \textbf{93.98\%} \\
 &  &  &  & \textbf{4} & \textbf{93.76\%} \\
 &  &  &  & \textbf{2} & \textbf{93.12\%} \\
\bottomrule[1.2pt]
\end{tabular}
\end{table*}

\subsection{Ablation Study}
\label{sec:ablation}
To further evaluate the performance of DPCNN, we conduct ablation experiments on four datasets. In the first set of experiments, we compare the performance of DPCNN and nonlinking DPCNN to evaluate the effect of coupling in DPCNN. In the second set of experiments, we aim to verify whether nonlinking DPCNN can achieve performance similar to LIFSNN by comparing them under the same network architecture. For simplicity, we fix $T$ at 4 and 10 for MNIST and N-MNIST datasets, respectively, and 6 for Fashion-MNIST and CIFAR-10 datasets. We set the decay factor of LIF neurons to 0.5, and the threshold to 1.
\begin{table}[htbp]
\renewcommand{\arraystretch}{1.2}
\centering
\caption{The accuracy of DPCNN, LIFSNN, and nonlinking DPCNN on four datasets}
\label{tab:ab-acc}
\begin{tabular}{cccc}
\toprule[1.2pt]
Dataset  & DPCNN & LIFSNN & Nonlinking DPCNN \\
\midrule[0.8pt]
MNIST         & 99.60\%          & 99.44\%           & 99.53\%                     \\
N-MNIST       & 99.54\%          & 99.45\%           & 99.47\%                     \\
Fashion-MNIST & 94.53\%          & 93.93\%           & 93.50\%                      \\
CIFAR-10      & 93.98\%          & 93.19\%           & 93.64\%      \\
\bottomrule[1.2pt]
\end{tabular}
\end{table}
\par
\begin{figure}[htbp]
	\centering
	\subfigure[MNIST]{\includegraphics[width=0.49\columnwidth]{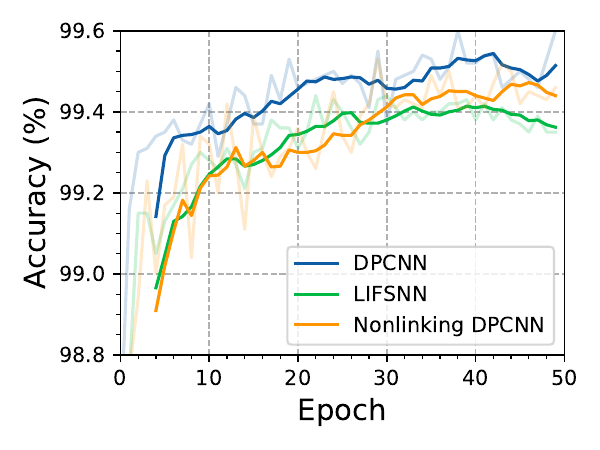}}  
	\subfigure[N-MNIST]{\includegraphics[width=0.49\columnwidth]{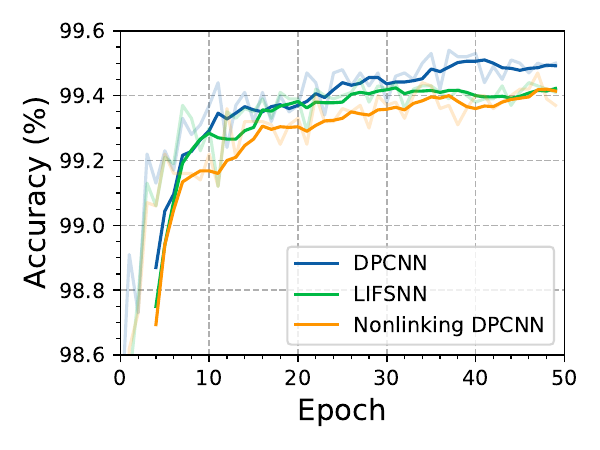}} 
    \subfigure[Fashion-MNIST]{\includegraphics[width=0.49\columnwidth]{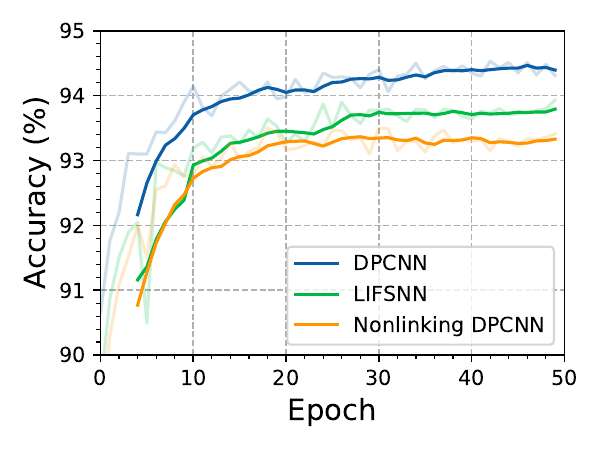}} 
	\subfigure[CIFAR-10]{\includegraphics[width=0.49\columnwidth]{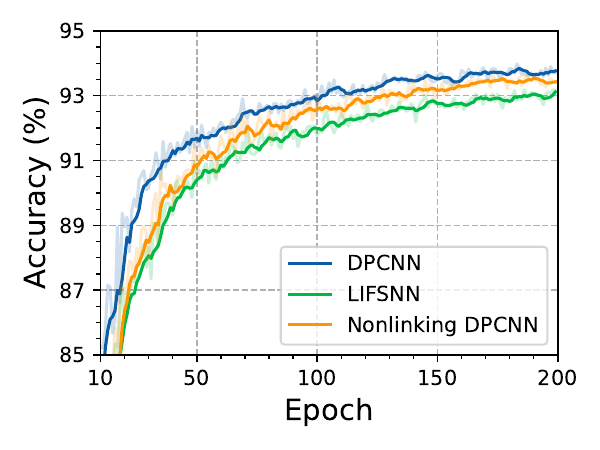}}  
	\caption{The test accuracy of DPCNN, LIFSNN, and nonlinking DPCNN on four datasets during training. The shaded curves represent the original data, while the solid curves are 64-epoch moving averages.}
    \label{fig:ab-acc}
\end{figure}


Tab. \ref{tab:ab-acc} presents the accuracy of DPCNN, LIFSNN, and nonlinking DPCNN on four datasets. The results indicate that DPCNN outperforms both LIFSNN and nonlinking DPCNN on all four datasets. While LIFSNN and nonlinking DPCNN exhibit similar performance, nonlinking DPCNN performs better than LIFSNN on MNIST, N-MNIST, and CIFAR-10 datasets. In contrast, LIFSNN achieves better results on the Fashion-MNIST dataset. Fig. \ref{fig:ab-acc} illustrates the test accuracy of the three models during training. It is evident that DPCNN converges faster and achieves better performance (blue curve).
\subsection{Analysis of Inter-Channel Coupling}
In this section, we will delve deeper into the properties of inter-channel coupling in DPCNNs.
\subsubsection{Inter-Channel Coupling Boosting performance efficiently}
\begin{figure}[htbp]
	\centering
	\includegraphics[width=0.8\columnwidth]{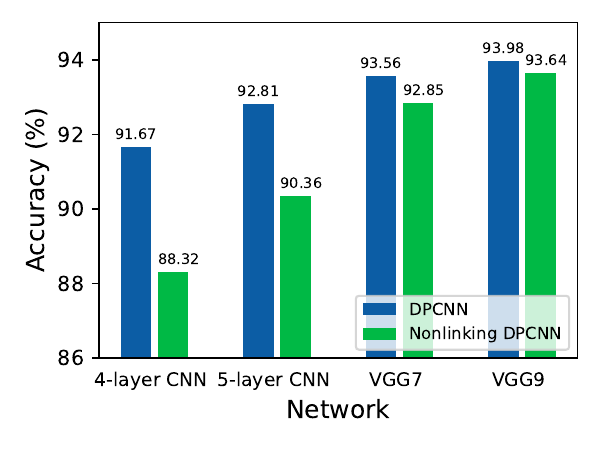}
	\caption{Performance improvement through inter-channel coupling at varying network depths.}
	\label{fig:net-acc}
\end{figure}

Fig. \ref{fig:net-acc} compares the accuracy of DPCNN and non-linking DPCNN on the CIFAR-10 dataset, using different network depths. The VGG7 architecture consists of 64C3-64C3-P2-128C3-128C3-128C3-P2-1024-10, the 5-layer CNN architecture is 64C3-P2-128C3-128C3-P2-1024-10, and the 4-layer CNN architecture is 64C3-P2-128C3-P2-1024-10. The results clearly indicate that inter-channel coupling significantly enhances the performance of all four networks. Moreover, the shallower the network, the greater the performance improvement from inter-channel coupling. Increasing network depth and introducing inter-channel coupling both effectively enhance performance, but the benefit of inter-channel coupling is that it does not require additional neurons. For example, the DPCNN based on the 4-layer CNN has 99.4K neurons with an accuracy of 91.67\%, while the non-linking DPCNN based on the 5-layer CNN has 132.1K neurons with an accuracy of 90.36\%.
\par
We further compare the performance improvement achieved by widening the network and introducing inter-channel coupling. To accomplish this, we will introduce widened versions of MNISTNet and VGG9, namely MNISTNetWide and VGG9Wide, respectively. In MNISTNetWide, the channel of convolutional layers is doubled, resulting in an architecture of 32C3-P2-32C3-P2-128-10. Similarly, the architecture of VGG9Wide is 64C3-64C3-P2-128C3-128C3-P2-256C3-256C3-256C3-P2-1024-10. 
\par
The experimental results presented in Tab. \ref{tab:wide} reveal that DPCNN achieves the highest accuracy on the Fashion-MNIST dataset, while on the CIFAR-10 dataset, it is surpassed only by non-linking DPCNN based on VGG9Wide, but outperforms LIFSNN based on VGG9Wide. These results suggest that inter-channel coupling can achieve compareble or even better performance compared to widening networks, which require additional neurons and synapses. The table also includes the computational costs of each corresponding network. It is evident that introducing inter-channel coupling can save a significant amount of computational resources compared to widening networks. This is because inter-channel coupling requires only a small number of additional synapses, while maintaining the same number of neurons, and thus requires less training time. For instance, compared to the LIF-based SNN with wide VGG9, DPCNN with VGG9 uses only 50\%, 53\%, and 73\% of neurons, synapses, and training time, respectively.

\begin{table*}[htbp]
\renewcommand{\arraystretch}{1.2}
\centering
\caption{The accuracy and computational costs of DPCNN, Nonlinking DPCNN, LIFSNN, and widened nonlinking DPCNN and LIFSNN on Fashion-MNIST and CIFAR-10}
\label{tab:wide}
\begin{tabular}{ccccccc}
\toprule[1.2pt]
Dataset & Network & Model & Accuracy & \#Neuron  & \#Synapse & Training time \\
\midrule[0.8pt]
\multirow{5}{*}{Fashion-MNIST} & \multirow{3}{*}{MNISTNet} & DPCNN & 94.53\% & 32.4K & 0.23M & 43s \\
 &  & Nonlinking DPCNN & 93.50\% & 32.4K & 0.21M & 26s \\
 &  & LIFSNN & 93.93\% & 32.4K & 0.21M & 26s \\ \cline{2-7}
 & \multirow{2}{*}{MNISTNetWide} & Nonlinking DPCNN &  94.05\%& 62.7K & 0.44M & 44s \\
 &  & LIFSNN &  94.12\%& 62.7K & 0.44M & 42s \\ \cline{1-7}
\multirow{5}{*}{CIFAR-10} & \multirow{3}{*}{VGG9} & DPCNN & 93.98\% & 246.8K & 8.06M & 175s \\
 &  & Nonlinking DPCNN & 93.64\% & 246.8K & 5.95M & 110s \\
 &  & LIFSNN & 93.19\% & 246.8K & 5.95M & 106s \\ \cline{2-7}
 & \multirow{2}{*}{VGG9Wide} & Nonlinking DPCNN & 94.30\% & 492.6K & 15.34M & 249s \\
 &  & LIFSNN & 93.55\% & 492.6K & 15.34M & 240s \\
\bottomrule[1.2pt]
\end{tabular}
\end{table*}

\begin{figure}[htbp]
	\centering
	\subfigure[Train accuracy on Fashion-MISNT]{\includegraphics[width=0.49\columnwidth]{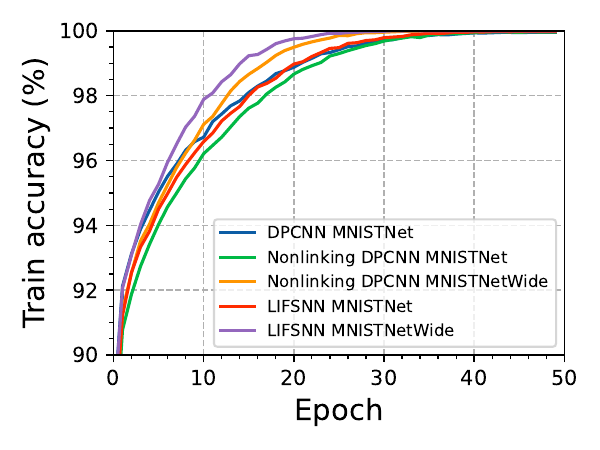}}
	\subfigure[Test accuracy on Fashion-MISNT]{\includegraphics[width=0.49\columnwidth]{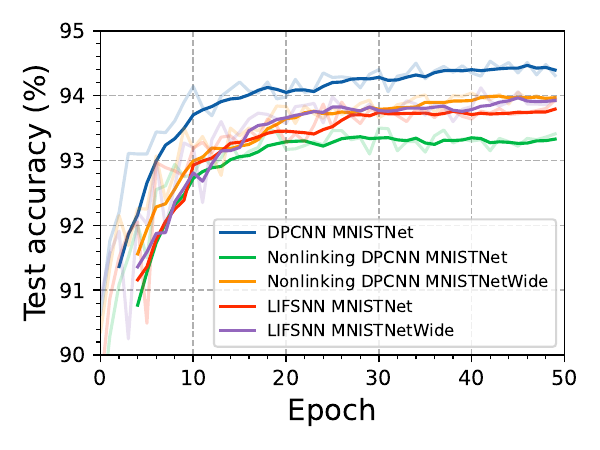}} 
    \subfigure[Train accuracy on CIFAR-10]{\includegraphics[width=0.49\columnwidth]{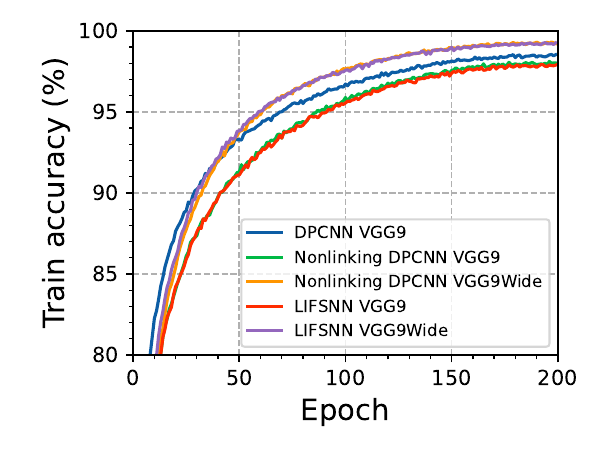}}
	\subfigure[Test accuracy on CIFAR-10]{\includegraphics[width=0.49\columnwidth]{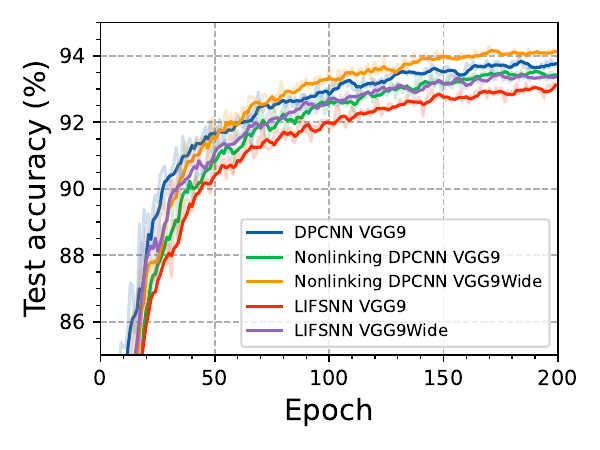}}  
	\caption{Performance comparison of DPCNN, nonlinking DPCNN, LIFSNN, and widened nonlinking DPCNN and LIFSNN. The test accuracy curves are smoothed with a moving average window of size 5.}
    \label{fig:wide-acc}
\end{figure}

Fig. \ref{fig:wide-acc} illustrates their learning curves. The results demonstrate that increasing the width of the networks can improve their training accuracy on both the Fashion-MNIST and CIFAR-10 datasets. However, wider networks may not necessarily result in better generalization performance due to the risk of overfitting.

\subsubsection{Multiplicative Coupling VS. Additive Coupling}
In PCNN neurons, the linking input modulates the feeding input through multiplication, which is referred to as multiplicative coupling in this work and is expressed as follows:
\begin{equation}
U = F(1+L)
\end{equation}
\par
Additive coupling is more common and can be expressed as:
\begin{equation}
U = F + L
\end{equation}
\par
We compared these two coupling mechanisms on the Fashion-MNIST and CIFAR-10 datasets, as shown in Tab. \ref{tab:coupling-add}. The results demonstrate that multiplicative coupling achieves better performance on both datasets, while additive coupling leads to a performance decrease on the CIFAR-10 dataset. One possible explanation for this is that the feeding input regulates the gradient of the coupling weights (i.e., linking synaptic weights), which may benefit the optimization process.
\begin{table}[htbp]
\renewcommand{\arraystretch}{1.2}
\centering
\caption{The accuracy on Fashion-MNIST and CIFAR-10 of intra-channel coupling and inter-channel coupling}
\label{tab:coupling-add}
\begin{tabular}{cccc}
\toprule[1.2pt]
Dataset& Without coupling & Multiplicative & Additive \\
\midrule[0.8pt]
Fashion-MNIST & 93.50\% & 94.20\%       & 94.53\%                           \\
CIFAR-10       & 93.64\%  & 93.12\%       & 93.98\%         \\    
\bottomrule[1.2pt]
\end{tabular}
\end{table}

\subsubsection{Intra-channel coupling VS. Inter-channel Coupling}
We compare the proposed inter-channel coupling with the previously used intra-channel coupling \cite{lian_overview_2021, cheng_lisnn_2020}. As shown in Table \ref{tab:intra-coupling}, we observed that intra-channel coupling slightly improves performance on the Fashion-MNIST dataset, but leads to a performance decrease on the CIFAR-10 dataset. This finding suggests that coupling between channels is essential for performance improvement. 
\begin{table}[htbp]
\renewcommand{\arraystretch}{1.2}
\centering
\caption{The accuracy on Fashion-MNIST and CIFAR-10 of intra-channel coupling and inter-channel coupling}
\label{tab:intra-coupling}
\begin{tabular}{cccc}
\toprule[1.2pt]
Dataset  & Without Coupling & Intra-channel & Inter-channel \\
\midrule[0.8pt]
Fashion-MNIST & 93.50\%        & 93.87\%                              & 94.53\%          \\
CIFAR-10      & 93.64\%      & 92.92\%                              & 93.98\%         \\
\bottomrule[1.2pt]
\end{tabular}
\end{table}

\subsubsection{Impact of Coupling scale}
The coupling scale is determined by the size of the convolution kernel. Tab. \ref{tab:coupling-scale} presents corresponding accuracies at different coupling scales. When using a $1\times1$ kernel, the accuracy on the CIFAR-10 dataset remains stable, while an improvement is observed on the Fashion-MNIST dataset. Using a $3\times3$ kernel leads to a significant accuracy improvement on both datasets. Despite a larger coupling scale, the accuracy of the $5\times5$ kernel is lower than that of the $3\times3$ kernel. The accuracy of the $3\times3$ kernel with dilation = 2, which has the same number of parameters and the same coupling scale as the $5\times5$ kernel, is also lower than that of the $3\times3$ and $5\times5$ kernels. These results indicate that a coupling scale of $3\times3$ is a suitable choice, and further increasing the coupling scale does not effectively improve performance.
\begin{table*}[htbp]
\renewcommand{\arraystretch}{1.2}
\caption{The impact of coupling scale on accuracies on Fashion-MNIST and CIFAR-10}
\label{tab:coupling-scale}
\centering
\begin{tabular}{cccccc}
\toprule[1.2pt]
\multirow{2}{*}{Dataset} & \multirow{2}{*}{Without coupling} & \multicolumn{4}{c}{Kernel size of coupling} \\ \cline{3-6}
 &  & $1\times1$ & $3\times3$ & $5\times5$ & $3\times3$, dilation = 2 \\ \midrule[0.8pt]

Fashion-MNIST & 93.50\% & 93.90\% & 94.53\% & 94.49\% & 94.45\% \\
CIFAR-10 & 93.64\% & 93.63\% & 93.98\% & 93.91\% & 93.57\% \\ 
\bottomrule[1.2pt]
\end{tabular}
\end{table*}

\subsection{Analysis of RFTD-BN}
\begin{figure}[htbp]
	\centering
	\subfigure[]{\includegraphics[width=0.49\columnwidth]{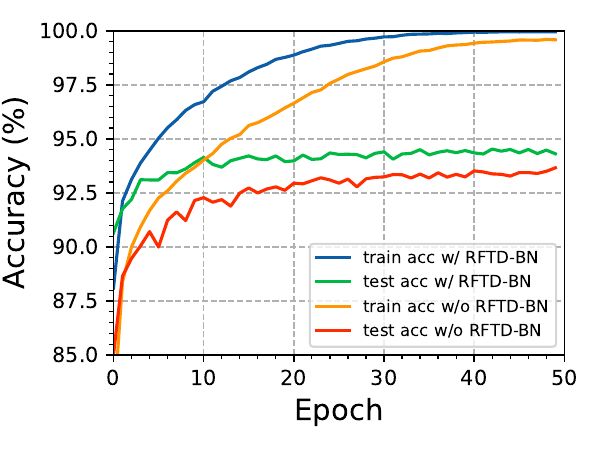}}
	\subfigure[]{\includegraphics[width=0.49\columnwidth]{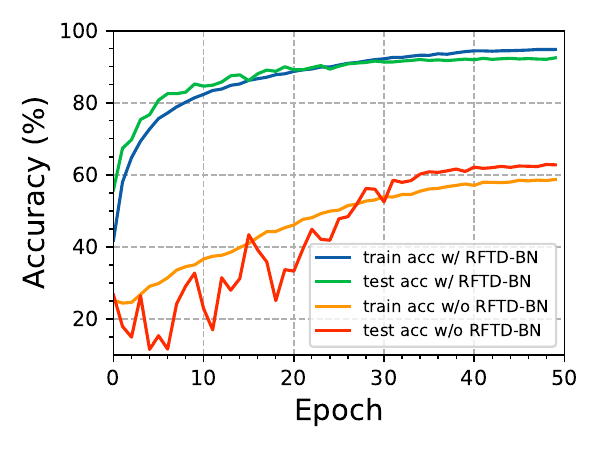}} 
	\caption{The train and test accuracy of DPCNN with and without RFTD-BN on (a) Fashion-MISNT with MNISTNet and (b) CIFAR-10 with VGG9.}
    \label{fig:rftd-bn}
\end{figure}

To capture the spatio-temporal characteristics of DPCNN, RFTD-BN utilizes separate sets of BN parameters for feeding and linking inputs at different time steps. To verify the necessity of RFTD-BN in DPCNN, we conduct ablation studies on Fashion-MNIST with MNISTNet and CIFAR-10 with VGG9. As shown in Fig. \ref{fig:rftd-bn}, in both cases, RFTD-BN accelerates convergence and improves accuracy. Notably, for deeper architectures (VGG9), DPCNN without RFTD-BN struggles to converge.
\par
We further introduce two additional BN methods, namely time-dependent batch normalization (TD-BN) and receptive field-dependent batch normalization (RFD-BN), for comparison.
\par
In TD-BN, we utilize the same BN parameters at time step $t$ to normalize both $\boldsymbol{I}^{t}_{F}$ and $\boldsymbol{I}^{t}_{F}$, as given by:
\begin{align}
    \hat{\boldsymbol{I}^{t}_{F}} = \mathrm{BN}^{t}(\boldsymbol{I}^{t}_{F}) &= \boldsymbol{\gamma}^{t} \odot \frac{\boldsymbol{I}^{t}_{F} - {\boldsymbol{\mu}}_{\mathcal{B},F}^{t}}{{\boldsymbol{\sigma}}_{\mathcal{B},F}^{t}} + \boldsymbol{\beta}^{t} \\ 
    \hat{\boldsymbol{I}^{t}_{L}} = \mathrm{BN}^{t}(\boldsymbol{I}^{t}_{L}) &= \boldsymbol{\gamma}^{t} \odot \frac{\boldsymbol{I}^{t}_{L} - {\boldsymbol{\mu}}_{\mathcal{B},L}^{t}}{{\boldsymbol{\sigma}}_{\mathcal{B},L}^{t}} + \boldsymbol{\beta}^{t} 
\end{align}
It is worth noting that this differs from TD-BN for nonlinking PCNN discussed in Sec. \ref{sec:rftd-bn}, where only the feeding input is considered.
\par
In RFD-BN, we use two separate BN layers to normalize $\boldsymbol{I}^{t}_{F}$ and $\boldsymbol{I}^{t}_{F}$, respectively. For each BN layer, the parameters are shared across time. The formula is as follows:
\begin{align}
    \hat{\boldsymbol{I}^{t}_{F}} = \mathrm{BN}_{F}(\boldsymbol{I}^{t}_{F}) &= \boldsymbol{\gamma}_{F} \odot \frac{\boldsymbol{I}^{t}_{F} - {\boldsymbol{\mu}}_{\mathcal{B},F}^{t}}{{\boldsymbol{\sigma}}_{\mathcal{B},F}^{t}} + \boldsymbol{\beta}_{F} \\ 
    \hat{\boldsymbol{I}^{t}_{L}} = \mathrm{BN}_{L}(\boldsymbol{I}^{t}_{L}) &= \boldsymbol{\gamma}_{L} \odot \frac{\boldsymbol{I}^{t}_{L} - {\boldsymbol{\mu}}_{\mathcal{B},L}^{t}}{{\boldsymbol{\sigma}}_{\mathcal{B},L}^{t}} + \boldsymbol{\beta}_{L} 
\end{align}
\par
\begin{figure}[htbp]
	\centering
	\subfigure[]{\includegraphics[width=0.49\columnwidth]{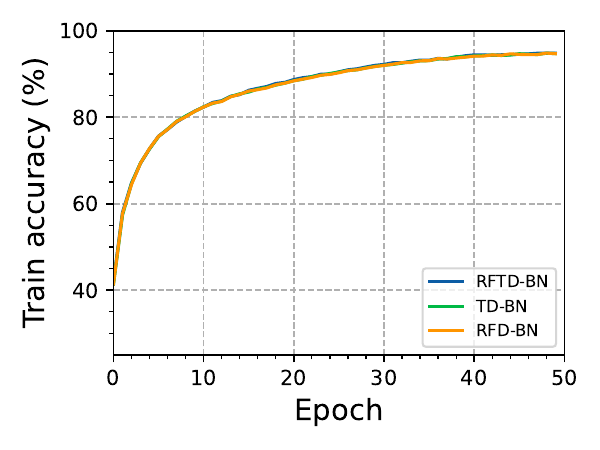}}
	\subfigure[]{\includegraphics[width=0.49\columnwidth]{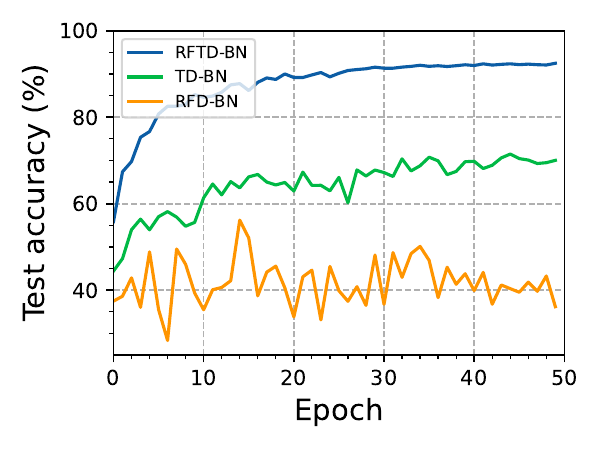}} 
	\caption{The train and test accuracy of DPCNN with/without RFTD-BN on (a) Fashion-MISNT with MNISTNet and (b) CIFAR-10 with VGG9.}
    \label{fig:bncifar}
\end{figure}

Fig. \ref{fig:bncifar} illustrate the train and test accuracy of DPCNN on CIFAR-10 with VGG9 under three different BN methods: RFTD-BN, RFD-BN, and TD-BN. We observe that all three BN methods can converge well on the train dataset with almost identical training accuracy. However, there is a significant difference in the test accuracy, where RFTD-BN can generalize well to the test dataset, while the other two methods exhibit lower testing accuracy. This suggests that separate statistics are required to capture the spatio-temporal distribution of spikes.

\subsection{Analysis of Hyperparameters}
\label{sec:hyperparameters}
\subsubsection{Impact of Leak Factor}
\begin{figure}[htbp]
	\centering
	\includegraphics[width=0.8\columnwidth]{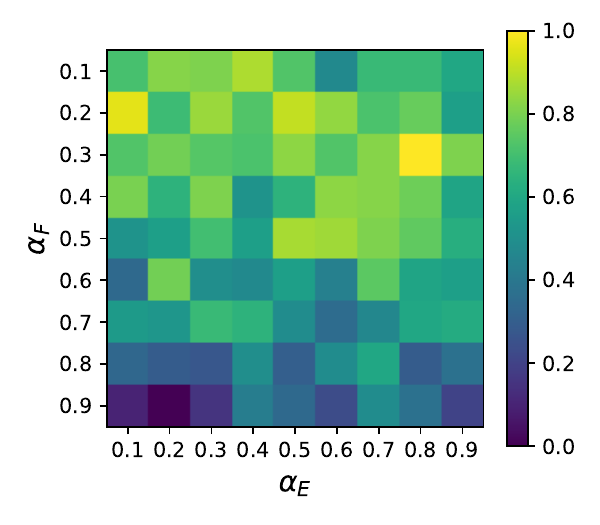}
	\caption{The impact of the leak factors $\alpha_{F}$ and $\alpha_{E}$ on accuracy on Fashion-MNIST.}
	\label{fig:heatmap}
\end{figure}

$\alpha_{F}$ and $\alpha_{E}$ control the decay rate of the feeding input and threshold. We investigate their impact on accuracy by performing a grid search on the Fashion-MNIST dataset. We set the search range for both parameters from 0.1 to 0.9, with a step size of 0.1. The results are presented in Fig. \ref{fig:heatmap}, where the accuracies are normalized to the range [0, 1]. The maximum, minimum, and average accuracies obtained are 94.72\%, 93.73\%, and 94.35\%, respectively, with a variance of 0.065\%. Notably, DPCNN demonstrates robustness to variations in $\alpha_{F}$ and $\alpha_{E}$ as it does not encounter any convergence issues.
\subsubsection{Impact of Time steps}
\begin{figure}[htbp]
	\centering
	\includegraphics[width=0.8\columnwidth]{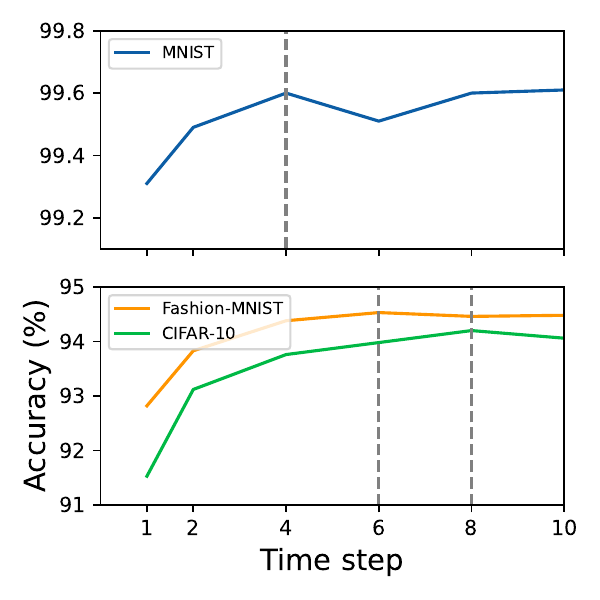}
	\caption{The impact of simulation time steps on accuracies on MNIST, Fashion-MNIST, and CIFAR-10}
	\label{fig:step-acc}
\end{figure}

We analyze the effect of the time step on accuracy using three static datasets, as depicted in Fig. \ref{fig:step-acc}. In general, as we increase the time steps, the accuracy improves. The dashed lines indicate the optimal time step required to achieve the highest accuracy, which are 4, 6, and 8 for MNIST, Fashion-MNIST, and CIFAR-10, respectively. This implies that as the dataset complexity increases, DPCNN requires a greater number of time steps.

\subsection{Robstness to Noise}
\label{sec:noise}
In this section, we will compare the robustness of DPCNN, nonlinking DPCNN, and LIFSNN.
\subsubsection{Robustness to Input Noise}
\begin{figure}[htbp]
	\centering
	\subfigure[]{\includegraphics[width=0.9\columnwidth]{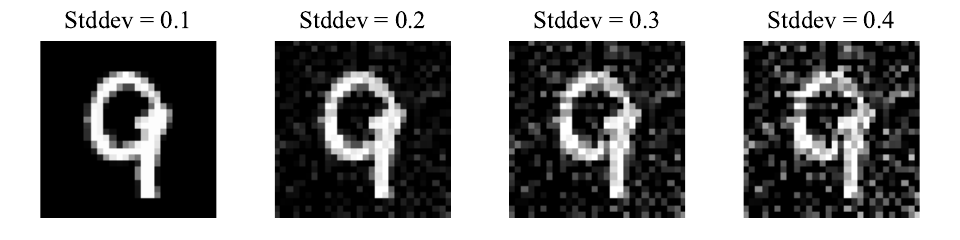}} 
	\subfigure[]{\includegraphics[width=0.75\columnwidth]{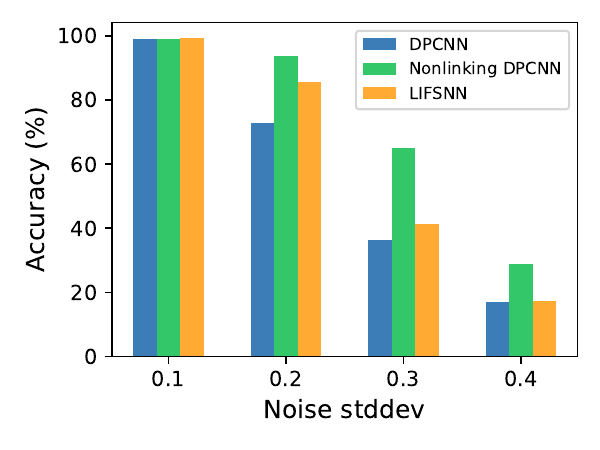}} 
	\caption{(a) A sample digit form MNIST with different levels of noise standard deviation. (b) Accuracy change on MNIST respect to the standard deviation of the gaussian noise. }
        \label{fig:noise}
\end{figure}

We investigate the robustness of DPCNN to Gaussian noise on the MNIST dataset. Throughout the experiments, the mean of the Gaussian noise is set to 0, and the noise intensity is controlled by adjusting the standard deviation of the noise distribution, as illustrated in Fig. \ref{fig:noise}(a). The accuracy results of DPCNN, nonlinking DPCNN, and LIFSNN on the noisy MNIST dataset are presented in Fig. \ref{fig:noise}(b). Notably, nonlinking DPCNN exhibits superior noise robustness, while DPCNN performs relatively poorly. This discrepancy can be attributed to the global propagation of local noise through coupling connections, which negatively impacts the performance of DPCNN.

\subsubsection{Robustness to Neuron Silencing Noise}
\begin{figure}[htbp]
	\centering
	\includegraphics[width=0.75\columnwidth]{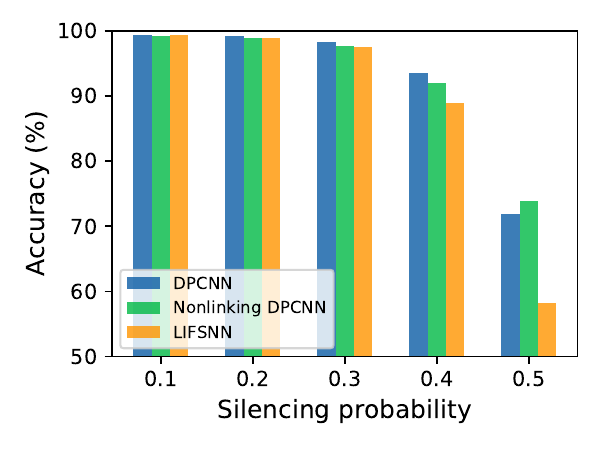}
	\caption{Accuracy change on MNIST respect to silencing probability.}
	\label{fig:nf}
\end{figure}

Neuron silencing refers to the situation where a portion of neurons fails to respond to stimuli \cite{yang2022training}. To simulate this phenomenon, we mask the output of neurons in the convolutional layers with a variable silencing rate ranging from 0.1 to 0.5. The results are presented in Fig. \ref{fig:nf}. When the failure probability is low, the performance of DPCNN, nonlinking DPCNN, and LIFSNN is comparable. However, at high silencing probabilities (e.g., 0.5), both DPCNN and nonlinking DPCNN demonstrate significant advantages.

\section{Conclusion}
In this work, we present deep pulse-coupled neural networks (DPCNNs) for image recognition tasks, which outperform current LIF-based SNNs on several mainstream datasets. To enhance DPCNN performance, we propose inter-channel coupling, which improves efficiency compared to simply widening networks. We also introduce receptive field and time dependent Batch Normalization (RFTD-BN), which is essential for training DPCNNs. By leveraging a visual cortex model, DPCNNs incorporate more bio-plausible mechanisms than existing SNNs. We believe that DPCNNs have the potential to advance the field of brain-like intelligence and deepen our understanding of how information is processed in the visual cortex.

\section*{Acknowledgments}
This work is supported by the Regional Project of the National Natural Science Foundation of China (Grant 82260364) and Natural Science Foundation of Gansu Province (Grants 21JR7RA510, 21JR7RA345)
\par 
Some experiments are supported by the Supercomputing Center of Lanzhou University. 
\par
Conflict of Interest: The authors declare that they have no conflicts of interest.



 

\bibliographystyle{IEEEtran}
\bibliography{refs.bib}

\end{document}